%% file: Romeo_VC_JWG_V01_Y.tex
\documentclass[twocolumn]{IEEEtran}

\usepackage{amsmath}
\usepackage{amsfonts}
\usepackage{amssymb}
\usepackage{graphicx}
\usepackage[latin1,ansinew]{inputenc}
\usepackage[latin1]{inputenc}
\usepackage{epstopdf}
\usepackage{psfrag}
\usepackage[hang]{subfigure}
\usepackage{color}

\newcommand{\images}{./}
\newcommand{\poincaresection}{\mathcal{S}}

\include{Romeo_VC_headerFile}

\begin{document}
\title{Self-synchronization and Self-stabilization of 3D Bipedal
Walking Gaits}
\author{Christine Chevallereau$^1$, Hamed Razavi$^2$, Damien Six$^1$, Yannick Aoustin$^1$, and Jessy Grizzle$^3$
\thanks{$^1$Christine Chevallereau, Damien Six and Yannick Aoustin are with Laboratoire des Sciences du Num\'erique de Nantes (LS2N), CNRS, Ecole Centrale de Nantes, Universit\'e de Nantes, Nantes, France, {\tt christine.chevallereau@ls2n.fr}
$^2$Hamed Razavi is with Biorobotics Laboratory (Biorob) of the \'Ecole Polytechnique F\'ed\'erale de Lausanne (EPFL), Lausanne, Switzerland,
        {\tt hamed.razavi@epfl.ch},
$^3$ Jessy W. Grizzle is with
Electrical Engineering and Computer Science Department of the University of Michigan, Ann Arbor, MI, USA, {\tt grizzle@eecs.umich.edu}}}%
\maketitle

\begin{abstract}
This paper seeks insight into stabilization mechanisms for periodic
walking gaits in 3D bipedal robots.
Based on this insight, a control strategy based on virtual
constraints, which imposes coordination between joints rather than a
temporal evolution, will be proposed for achieving asymptotic convergence toward a periodic motion. For planar bipeds with one degree of underactuation, it is known that a
vertical displacement of the center of mass---with downward velocity at the step transition---induces stability of a walking gait. This
paper concerns the \textit{qualitative extension} of this type of property to 3D walking {\color{black} with two degrees of underactuation}.
It is shown that a condition on the position of the center of mass in the horizontal plane at the transition between steps induces synchronization between the motions in the sagittal and frontal planes. A combination of the conditions for self-synchronization and vertical oscillations leads to stable gaits. The algorithm for self-stabilization of 3D walking gaits is first developed for a simplified model of a walking robot (an inverted pendulum with variable length legs), and then it is extended to a complex model of the humanoid robot Romeo using
the notion of Hybrid Zero Dynamics. Simulations of the model of the robot illustrate the efficacy of the method and its robustness.
\end{abstract}


%

\section{Introduction}

Despite a growing list of bipedal robots that are able to walk in a
laboratory environment or even outdoors, stability mechanisms for 3D
bipedal locomotion remain poorly understood. It would be very
satisfying to be able to point at a robot and say, ``it can execute
an asymptotically stable walking gait, and the stability is achieved
through such and such theorem, principle, method, etc.'' For planar
(aka 2D) robots with one degree of underactuation, virtual
constraints and hybrid zero dynamics provide an integrated gait and
controller design method that comes with a formal closed-form stability
guarantee \cite{Westervelt2007a}[pp.~128-135]. Moreover, the stability condition
is physically meaningful: the velocity of the center of mass (CoM)
at the end of the single support phase must be directed downward \cite{Chevallereau}.

For fully actuated 3D bipeds and for 3D bipeds with a special form of
underactuation, the 2D results extend nicely
\cite{Abba2016,Nijmeijer2013}. But for 3D robots with more than one
degree of underactuation, not much is known.

{\color{black} Basing the control law design of a fully actuated robot on a model with either a passive ankle or a point foot contact is an interesting intermediate view of the robot. Even in case
of a fully actuated robot, the ankle torques in the frontal and
sagittal planes are limited by the size of the foot. A model
with two degrees of underactuation is therefore useful in order to
avoid the use of these torques for a nominal (unperturbed) gait. This allows the limited ankle
torque to be saved for adapting the foot orientation
in case of uneven terrain or for increasing the robustness of
the control strategy in the face of perturbations.}

{\color{black} The possibility and interest of extending the method of virtual
constraints to robot models with two degrees of
underactuation have been shown in \cite{Chevallereau2008,Razavi2016,AkBuGri2015,AkGri2016,GriGri2016}.
While the implementation of the control strategy is quite
straightforward, the choice of the virtual constraints is not
obvious. Their selection is often based on an optimization process
and choice of appropriate controlled outputs and/or the
introduction of a high-level event-based control strategy, neither of which provides insight into the
stability mechanism. The
objective of this paper is to provide some qualitative results
on gait characteristics that \textit{when used to define virtual
constraints} lead to asymptotically stable walking.}


In contrast to fully actuated bipedal walking, passive walking
\cite{Mgeer90a, MGeer90b} can be seen as an emergent behavior, where
alternating leg impacts of a biped and the pull of gravity combine
to produce asymptotically stable motions on mild downward slopes.
The obtained gait is very efficient from an energy consumption point
of view and can be extended to walking on flat ground \cite{spong,
collins, wisse}. One shortcoming of such an approach is limited
robustness with respect to perturbations.

As there is no actuation in passive robots, the obtained walking
gaits may be called self-stable.  The self-stability property is well
understood in 2D walking gaits \cite{Chevallereau2003, PoMaAmAm2016}, however,
little is known on why a passive robot can demonstrate stable
walking in 3D.  Only a few
studies have been devoted to the investigation of analytical properties that
lead to asymptotically stable gaits in 3D underactuated robots
\cite{theseMIT, Razavi2016}. In \cite{Chevallereau2008} for example,
it has been shown that when a controller design method that yielded
asymptotically stable gaits in planar underactuated walking
\cite{Chevallereau2003} is applied to 3D walking, unstable solutions are common.

The notion of \textit{self-synchronization},
introduced in \cite{Razavi2014} and generalized in
\cite{Razavi2016}, sheds some light on the stability mechanisms in
3D legged locomotion.

The aim of the paper is to provide insight towards understanding the
mechanisms of asymptotic stability in periodic walking of 3D
underactuated robots. Using a control law based on virtual constraints for 3D bipedal robots with two degrees of underactuation, our objective is to propose virtual constraints leading to self stabilization.
In other words, without the use of
event-based control, the obtained gaits are
asymptotically stable.

\begin{figure}
\centering
        \includegraphics[width=0.4 \linewidth]{\images 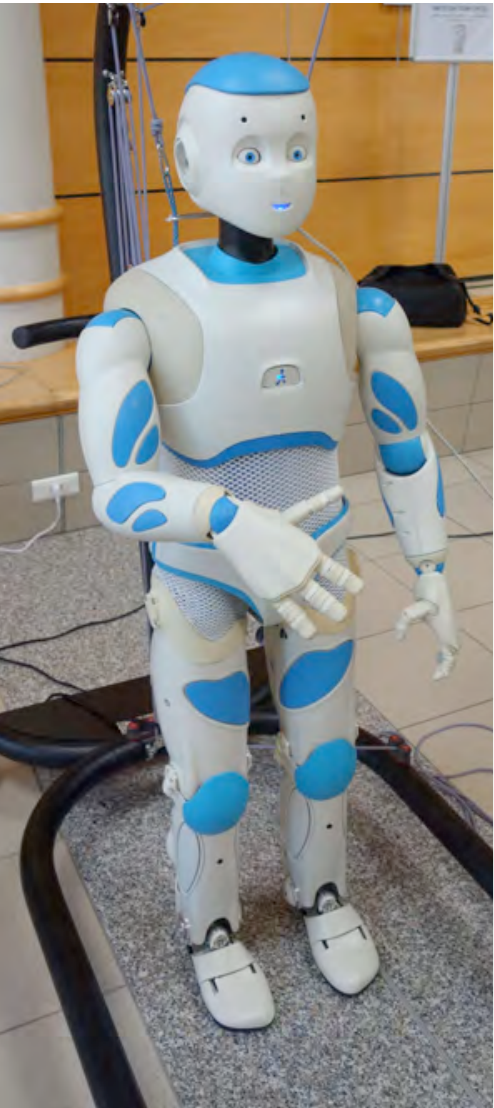}
        \caption{The humanoid robot Romeo developed by Aldebaran robotics.}
        \label{fig:romeo}
\end{figure}

As opposed to the formal theorems proven in  \cite{Westervelt2007a} for 2D robots, and for the self-synchronization of the 3D LIP \cite{Razavi2014, Razavi2016} the results here for 3D will be numerical in nature. By normalizing the studied model, qualitative gait features leading to asymptotically stable periodic orbits will be uncovered, nevertheless.
The method will be first demonstrated on an inverted pendulum model
and then will be extended to a full model of the humanoid Romeo
\cite[Chapter~7]{razavi2016thesis}, shown in Figure \ref{fig:romeo}.

{\color{black} The remainder of the paper is organized as follows. In Section \ref{sec:background}, as background, the control method based on virtual constraints for a 3D humanoid robot with 2 degrees of underactuation is recalled. The stability of the full-order model of the robot is discussed in relation to an induced reduced-order model, called the Hybrid Zero Dynamics (HZD). In Section \ref{sec:Motivation}, the difficulty of choosing appropriate virtual constraints is highlighted. Moreover, an important difference between planar and 3D walking, which is due to an increase of the degrees of underactuation, is illustrated. In particular, it is shown that unlike the planar case, where at the impact the set describing the final configuration of the robot in the reduced model consists of a single point, in 3D walking with two degrees of underactuation, this set is a manifold of dimension one. In Section \ref{sec:VLIP}, where we study a very simplified model of a humanoid robot, the role of this subset in achieving self-synchronization for the Linear Inverted Pendulum model (LIP) will be demonstrated. Moreover, a new geometric interpretation of the self-synchronization property will be provided in this section.} Subsequently, in Section \ref{sec:VLIP-stab}, we
introduce vertical oscillations of the CoM for the pendulum model.
\textit{Self-stabilization} properties will be numerically studied
in Section \ref{sec:stability_res}. In Section \ref{sec:humanoid}, the results obtained on the simplified model will be
extended to a realistic model of a humanoid robot. A numerical study shows how
the stability properties obtained for the pendulum extend to a
complete model of a bipedal robot. Simulation results illustrate the efficiency of the approach.

{\color{black}
\section{Background: A Reduced-order Model Associated with Underactuation}
\label{sec:background}
This section reviews how to create a reduced-order model within the full-dimensional hybrid model of a bipedal robot with the following properties:
 \begin{itemize}
 \item the dimension of the reduced-order model is determined by the number of degrees of underactuation of the robot;

 \item the periodic orbits of the reduced-order model are periodic orbits of the robot;

 \item locally exponentially stable periodic orbits in the reduced-order model can always be rendered locally exponentially stable in the full-order model of the robot through the use of feedback control; and

\item no approximations are being made.
 \end{itemize}
 Based on the above properties, it can be seen that the reduced-order model captures the effects of underactuation on the design of stabilizing feedback control laws for bipedal robots. Later sections of the paper develop properties of this reduced-order model that lead to existence of locally exponentially stable walking gaits.}

 \subsection{Full-order hybrid model}
{\color{black}
Let $q \in {\cal Q}$ denote the generalized coordinates for a humanoid robot in single support (one foot on the ground) and assume the stance ankle has two DoFs (pitch and roll), both assumed to be passive, that is, unactuated. Typically, $q$ consists of body coordinates, such as joint angles of the stance leg, swing leg, upper-body, and in our case, two additional world-frame coordinates capturing the pitch and roll degrees of freedom in the ankle. The robot is assumed to have right/left symmetry and the gait studied corresponds to walking along a straight line in the sagittal plane. Moreover, the gait is assumed to be composed of single support phases separated by impacts, which are modeled as instantaneous changes of support. At each transition, a relabeling of the joint variables is introduced. This relabeling allows us to work with a single model of the robot.

The Lagrangian is assumed to have the form $$L(q,\dot{q})=\frac{1}{2} \dot{q}^T D(q) \dot{q}- V(q),$$
and the single support model is given by the standard Lagrange equations
\begin{equation}\label{eq:lagrangian}
\frac{\text{d}}{\text{d}t}\frac{\partial L(q,
{\dot{q}})}{\partial\dot{q}}-\frac{\partial
L(q,{\dot{q}})}{\partial q}=B u,
\end{equation}
where the $N \times (N-2)$ torque distribution matrix $B $ is constant and has rank $N-2$, and
$u$ is an $(N-2)$-dimensional vector associated to the torques of the
joint coordinates.

Equation \eqref{eq:lagrangian}
can also be written in the form
\begin{equation}\label{eq:mod_dyn}
D(q) \ddot q +H(q,\dot q)=Bu,
\end{equation}
where $D(q)$ is the (positive definite) inertia matrix, and $ H(q,\dot q)$ groups the centrifugal, Coriolis and gravity terms.
Letting $x=[q^\top, \dot{q}^\top]^\top$, standard calculations lead to a state-variable model
\begin{equation}
\label{eqn:affine}
  \dot x  = f(x) + g(x) u,
  \end{equation}
which is affine in the control torques.

For simplicity, impacts are assumed to occur with the swing foot parallel to the walking surface. Let
\begin{equation}
  \label{eqn:grizzle:modeling:S}
  \poincaresection := \{ x~|~z_s = 0,\; \dot{z}_s <0 \},
\end{equation}
where $z_s$ is the height of the swing foot above the walking surface.  It is assumed that $ \poincaresection$ is a smooth $(2N-1)$-dimensional manifold in the state space of the robot. The widely used impact model of Hurmuzlu \cite{Hurmuzlu1994} leads to an algebraic representation of the jumps in the velocity coordinates when the swing foot contacts the ground, and hence to a hybrid model of the robot:
 \begin{equation}\label{eqn:full_hybrid_model_walking}
  \begin{array}{llll}
    \dot x & = f(x) + g(x) u,   & x^- &\notin \poincaresection,\\
    x^+ & = \impactmap(x^-), & x^- &\in \poincaresection,
  \end{array}
\end{equation}
where $x^{+}:=\lim_{\tau \searrow t} x(\tau)$  (resp. $x^{-}:=\lim_{\tau \nearrow t} x(\tau)$) is the state value just after (resp. just before) impact. The  control-affine ordinary differential equation (ODE) describes the dynamics of the swing phase, whereas the algebraic equation describes the impacts of the swing foot with the ground.
 }
 {\color{black} This algebraic equation can be decomposed into as follows:
  \begin{equation}
  \begin{array}{ll}
      q^+ & = \impactmappos(q),\\
    \dot q^+ & =\impactmapvel(q) \dot q^-.
  \end{array}\label{eq:impact}
\end{equation}}
{\color{black} The first matrix equation of \eqref{eq:impact} describes the relabelling of the coordinates at impact, while the second define the jump velocity coordinates.}

\subsection{Virtual constraints and an exact reduced-order model}

{\color{black} It is well understood in mechanics that a set of (regular) holonomic constraints applied to a Lagrangian model of a mechanical system leads to another Lagrangian model that has dimension equal to the dimension of the original model minus twice the number of independent constraints. In the case of mechanical systems, the constraints are imposed by a vector of generalized forces (aka Lagrange multipliers) that can be computed through the principle of virtual work.  When computing the resulting reduced-order model, no approximations are involved, and solutions of the reduced-order model are solutions of the original model along with the inputs arising from the Lagrange multipliers.

Virtual holonomic constraints are functional relations on the configuration variables of a robot's model that are achieved through the action of the robot's joint torques and feedback control instead of physical contact forces. They are called \textit{virtual} because they can be re-programmed on the fly in the control software. Like physical constraints, under certain regularity conditions, virtual constraints induce a low-dimensional invariant model called the \textit{zero dynamics} \cite{BYRNESC91,WesGriKo2003,Westervelt2007a}. A detailed comparison of  ``physical constraints'' versus ``virtual constraints'' is provided in \cite{ArXivpaper}. In the following, a brief overview is given.


A total of $N-2$ virtual constraints can be generated by the $N-2$ actuators. The virtual constraints are first expressed as outputs applied to the model \eqref{eqn:affine}, and then a feedback controller is designed that  asymptotically drives them  to zero. One such controller, based on the computed torque, is given in the Appendix; others can be found in \cite{AGSG_TAC_2014}. Here, the virtual constraints will be expressed in the form
\begin{equation}\label{eqn:grizzle:FdbkDesignApproach:output_fcn_structure_new_coord}
  y = h({q}) := \aq-h_d(\uq),
\end{equation}
where $\aq \in \Q_{\rm c} \subset \mathbb{R}^{(N-2)}$,
and $\uq \in \Q_{\ulbl} \subset \mathbb{R}^{2}$ with $(\aq, \uq)$ forming  a set of generalized configuration variables for the robot, that is, such that a diffeomorphism
$$F:\Q_{\rm c} \times  \Q_{\ulbl} \to \Q$$
exists. The variables $\aq$ represent physical quantities that one wishes to ``control'' or ``regulate'', while the variables $\uq$ remain ``free''. Later, a special case of $h_d(\uq)$ will be employed based on a \textit{gait phasing variable}, which makes it easier to interpret the virtual constraints in many instances.

When the virtual constraints are satisfied,  the relation $ \aq=h_d(\uq)$ leads to the mapping
 $F_c:  \Q_{\ulbl} \to \Q$  by
 \begin{equation}
 \label{eqn:virtualConstraintInducedMapping}
F_c(\qfree):= F(h_d(\qfree),\qfree)
\end{equation}
being an embedding; moreover, its image defines a constraint manifold in the configuration space, namely
\begin{equation}
\label{eqn:grizzle:ConstraintManifoldInq1q2}
\tilde{\Q}=\left\{q \in \Q~|~ q = F_c(\qfree),~\qfree \in Q_{\rm f}  \right\}.
\end{equation}
The constraint surface for the virtual constraints is
\begin{align}
\label{eqn:grizzle:VC:ZeroDynManifold}
\zdmanifold:&=\{(q,\dot{q}) \in T\Q~| y = h(q) = 0, ~\dot{y}=\frac{\partial h(q)}{\partial q}\dot{q}=0\} \nonumber \\
&=\{(q,\dot{q}) \in T\Q~| q=F_c(\uq), \dot{q}=J_c(\uq) \duq, ~~(\uq, \duq) \in T\Q_{\ulbl} \},
\end{align}
where
\begin{equation}
J_c(\uq) :=\frac{\partial F_c(\uq)}{\partial \uq}.
\end{equation}
The terminology ``zero dynamics manifold'' comes from \cite{BYRNESC91}. It is the state space for the internal dynamics compatible with the outputs being identically zero.  For the underactuated systems studied here, the dimension of the zero dynamics manifold is $4$, due to the assumption of two degrees of underactuation.

Let $B^\perp$ be the left annihilator of $B$, that is, a $2 \times N$ matrix of rank $2$ such that $B^\perp B=0$.
Lagrange's equation \eqref{eq:lagrangian} then gives
$$ \frac{\text d}{\text dt} \big( B^\perp \frac{\partial L}{\partial \dot{q} }\big) =  B^\perp  \frac{\partial L}{\partial q },$$
because $B^\perp B u=0$. The term $B^\perp \frac{\partial L}{\partial \dot{q} }$ is a form of \textit{generalized angular momentum}.

Restricting the full-system to the the zero dynamics manifold, gives
$$  \dusigma  = \kappa(\uq,\duq),$$
where
\begin{align} \usigma&:=M(\uq) \duq \nonumber \\
\kappa(\uq,\duq) &:= \left. B^\perp  \frac{\partial L(q,\dot{q}) }{\partial q } \right|_{
\begin{array}{l} q=F_c(\uq)\\ \dot{q}=J_c(\uq) \dot{\uq}\end{array}
}. \nonumber
\end{align}
The invertibility of $M(\uq)$ is established in \cite{ArXivpaper} under the assumption that the decoupling matrix used in \eqref{eq:appendix} is full rank.

$$z=\left[\begin{array}{c} z_1\\ z_2 \end{array}\right]:=\left[\begin{array}{c} \uq\\ \usigma \end{array}\right]$$
gives
\begin{align}
  \dot z & =    \left[
      \begin{array}{c}
        M^{-1}(z_1) z_2\\
        \bar{\kappa}(z_1,z_2)
      \end{array}\right] \label{eqn:CC:modeling:model_zeroAlternative}\\
  & =: f_{zero}(z),   \nonumber
\end{align}
where $\bar{\kappa}(z_1,z_2) = \kappa(z_1, M^{-1}(z_1)z_2)$.  The properties of this four-dimensional system are determined by the Lagrangain dynamics of the full-order model and the choice of the virtual constraints through the mapping \eqref{eqn:virtualConstraintInducedMapping}.
The form of the model is similar to the one used to analyze planar systems with one degree of underactuation \cite[Remark~5.2]{Westervelt2007a}.
 \\

}

\subsection{Reduced-order hybrid model (hybrid zero dynamics)}

{\color{black}
Analogous to the full-order hybrid model \eqref{eqn:full_hybrid_model_walking} is the so-called hybrid zero dynamics in which the zero dynamics manifold must satisfy
\begin{equation}\label{eqn:grizzle:impact_invar_cond}
  \impactmap (\poincaresection \cap \zdmanifold) \subset \zdmanifold.
\end{equation}
This condition means that when a solution evolving on $\zdmanifold$ meets the switching surface, $\poincaresection$, the new initial condition arising from the impact map is once again on $\zdmanifold$.


The invariance condition \eqref{eqn:grizzle:impact_invar_cond} is equivalent to
\begin{align}
0=&h\circ \impactmappos(q) \\
0=& \left. \frac{\partial h(\bar{q})}{\partial q} \right|_{\bar{q}=\impactmappos(q)} \impactmapvel(q)\, \dot{q}
\end{align}
for all $(q;\dot{q})$ satisfying
\begin{align}
 h(q) =0, ~~\frac{\partial h(q)}{\partial q} \dot{q} =& 0 \\
z_s(q) = 0,~~\dot{z}_s(q, \dot{q})<& 0.
\end{align}
At first glance, these conditions may appear to be difficult to meet. In the case of models with one degree of underactuation, however, it is known that if a single non-trivial solution of the zero dynamics satisfies these conditions, then all solutions of the zero dynamics will satisfy them \cite[Thm.~5.2]{Westervelt2007a}. In the case of systems with more than one degree of underactuation, systematic methods have been developed which modify the virtual constraints ``at the boundary'' and allow the conditions to be met \cite{MoGri2009}. Very straightforward implementations of the result are presented in a robotics context in \cite{CHGRSHIH09} and \cite{GrGr2015ACC}, and we refer the reader to these paper for the details.

With the invariance condition of the impact map, the reduced order hybrid system, called \textit{hybrid zero dynamics}(HZD), can be written in the form
\begin{equation}\label{eqn:grizzle:zero_dynamics:reduced_hybrid_model}
  \Sigma_\zero: \begin{cases}
    \begin{aligned}
      \dot z & = \fzero(z), &    z^- & \notin \poincaresection\cap \zdmanifold\\
      z^+ & = \impactmapzero(z^-), & z^- & \in \poincaresection\cap \zdmanifold,
    \end{aligned}
    \end{cases}
\end{equation}
with $\impactmapzero:= \left. \impactmap \right|_{\poincaresection\cap \zdmanifold}$. It is proven in \cite{Westervelt2007a} that solutions of \eqref{eqn:grizzle:zero_dynamics:reduced_hybrid_model} lift to solutions of \eqref{eqn:full_hybrid_model_walking}; indeed, \eqref{eqn:virtualConstraintInducedMapping} results in
\begin{align*}
q(t)&= F_c(z_1(t))\\
\dot{q}(t) &= J_c(z_1(t)) \dot{z}_1(t)\\
&= J_c(z_1(t)) M^{-1}(z_1(t)) z_2(t).
\end{align*}
It is also proven in \cite{Westervelt2007a} that locally exponentially stable periodic solutions of \eqref{eqn:grizzle:zero_dynamics:reduced_hybrid_model} lift to solutions of \eqref{eqn:full_hybrid_model_walking}, which can be locally exponentially stabilized.
\vspace{.1cm}

\noindent \textbf{Remark 1:} Low-dimensional pendulum models are  \textit{approximate} representation of the the swing phase dynamics of a robot, and normally when they are used, the impact map is ignored. The zero dynamics is an \textit{exact} low-dimensional model that captures the underactuated nature of the robot, and moreover, the hybrid zero dynamics is an  \textit{exact} low-dimensional ``subsystem'' of the full-order hybrid model that captures the underactuated dynamics of the hybrid model.
}


{\color{black}
\section{Self-stability and the HZD}
\label{sec:Motivation}


The use of the virtual constraints allows the stability of the full-order model to be studied on the HZD. For a system of two degrees of underactuation, the HZD is reduced to a system of dimension four for any number of actuated joints. The associated Poincar\'e map has dimension three. This paper seeks conditions for the eigenvalues of the Jacobian of the Poincar\'e map to lie within the unit circle, thereby assuring local exponential stability of a periodic orbit. More precisely, this paper seeks guidelines for the selection of the virtual constraints so that eigenvalues of the Jacobian lie within the unit circle.

Once the virtual constraints are defined, we know how to write a control law that drives them to zero and the walking gait will be stable or not depending on the stability of the HZD. If the chosen virtual constraints induce stable walking, we will call this \emph{self-stability}, in analogy with passive walkers.}

{\color{black}
\subsection{Choosing the virtual constraints by physical intuition}

Choosing the virtual constraints \eqref{eqn:grizzle:FdbkDesignApproach:output_fcn_structure_new_coord} means not only selecting $h_d(\uq)$, but also $\uq$ and $\aq$. One objective of this work is to give a physical interpretation of stability of walking; thus, we choose a set of variables $\aq$, $\uq$ that are physically meaningful. Several choices seem possible. Since the effect of gravity appears to be crucial and the stance ankle is supposed to be unactuated in the sagittal and frontal planes, the free variables are chosen as the position of the CoM in the horizontal plane  $x_M,~y_M$. Moreover, to make our results independent of the step width $D$ and length $S$, we consider normalized variables, namely, $\uq=(X,Y)^\top$,  $X=\frac{x_M}{S}$ and $Y=\frac{y_M}{D}$.

Concerning the controlled variables, in principle, we wish to use meaningful coordinates that appear in both models that we will study in the next sections, that is, a simplified inverted pendulum model and a realistic model of Romeo. Since the horizontal position of the CoM is used in $\uq$, it makes sense to consider the vertical position $z_M$ as one of the coordinates. Moreover, we also include the Cartesian position of the swing foot tip as a part of the controlled variables because these variables have a significant contribution to the change of support. Hence, for the simplified model of Section \ref{sec:VLIP}, the set of controlled variables will be $\aq = (z_M, X_s, Y_s, z_s)$ {\color{black} where $z_s$ is the vertical position of the swing foot and  $X_s, Y_s$ are normalized horizontal positions of the swing foot}. For the complete model of Romeo and other realistic bipeds, $\aq$ will consist of the variables just listed for the simplified model, together with a minimal set of body coordinates to account for the remaining actuators on the robot. For Romeo, this will mean adding the orientation of the swing foot, and remaining joints in the torso and stance leg as described in Section \ref{sec:humanoid}.}

{\color{black}
\subsection{From one degree of underactuation to two degrees of underactuation}

By definition, the condition of transition from one step to the next is defined by \eqref{eqn:grizzle:modeling:S}. In this condition, the vertical velocity of the foot being negative at impact ensures a well-defined transversal intersection of the solution with the transition set. In the following discussion, we compare the set of possible configurations of the robot at the change of support in the case of one degree of underactuation versus two degrees of underactuation. 

In the context of the virtual constraints, the evolution of the swing leg is imposed as a function of $\uq$. Thus, the switching condition can be viewed as a condition on the free configuration $\uq$. We will define a \textit{switching configuration manifold} $\mathbb{S}$: 
\begin{equation}
\label{eq:switching_manifold}
\mathbb{S}=\left\lbrace \uq|z_s^d(\uq)=0\right\rbrace.
\end{equation}
where $z_s^d$ is the virtual constraint for the height of the swing leg.

In the case of one degree of underactuation, the switching configuration manifold contains two points corresponding to the final value $\uq^-$ of the periodic motion and the initial one $\uq^+$, and a proper impact only requires $\dot{z} <0$ at the end of the step. 
For planar walking in the sagittal plane \cite{Westervelt2007a}, the stability condition is determined completely by the periodic motion itself, in other words, it is independent of the choice of virtual constraints that induce it. Moreover, the stability condition is physically meaningful: the velocity of the CoM
at the end of the single support phase (for the periodic motion) must be directed downward \cite{Chevallereau}.

In the case of two degrees of underactuation, the manifold \eqref{eq:switching_manifold} depends on at least two free variables $X$ and $Y$, and therefore under mild regularity conditions it includes an infinite number of points. Hence, the switching configuration manifold of the robot right before the impact is represented by the solutions of
\begin{equation}
z_s^d(\uq)= z_s^d(X,Y) = 0.
\label{eq:switching_2}
\end{equation}
Moreover, for two degrees of underactuation, it has been shown that a given periodic gait can be asymptotically stable or unstable depending on the choice of the virtual constraints used to induce the motion \cite{Chevallereau2008}. We want to understand how this happens.

As a first observation, the choice of the virtual constraint, $z_s^d$, clearly affects the set of free variables belonging to the \emph{switching manifold}:
\begin{equation}\label{eq:switching}
\mathbb{S}=\left\lbrace (X,Y)|z_s^d(X,Y)=0\right\rbrace.
\end{equation}
We will show in this paper that the shape of the switching manifold \eqref{eq:switching} has a crucial effect on the stability of walking. In particular, we show that
\begin{itemize}
\item in the case of the Linear Inverted Pendulum (LIP), a very common simplified  model for a humanoid robot, a condition to obtain self-synchronization between the model's motions in the lateral and sagittal planes is obtained as a condition on the tangent of $\mathbb{S}$ at the point where it is intersected by the periodic orbit; and

\item when the LIP model is extended to an inverted pendulum model with a stance leg of variable length, self-synchronization plus a downward motion of the CoM yields self-stability of a periodic motion.
\end{itemize}

The results are then extended to a realistic model of the humanoid Romeo and illustrated in simulation.}


%
\section{Simple model: Variable Length Inverted Pendulum}
\label{sec:VLIP}%

Two of the main features involved in dynamic walking are the
role of gravity and the limited torque available at the stance ankle before
the foot rolls about one of its edges.
Emphasizing these aspects, the inverted pendulum model has been used
in numerous humanoid walking control applications, such as
\cite{Kajita2001}, \cite{Doi2005}, and \cite{Koolen2012}.

\subsection{Modeling of a Variable Length Inverted Pendulum}
\label{sec:VLIP-mod}
\begin{figure}
\centering
        \includegraphics[width=0.4 \linewidth]{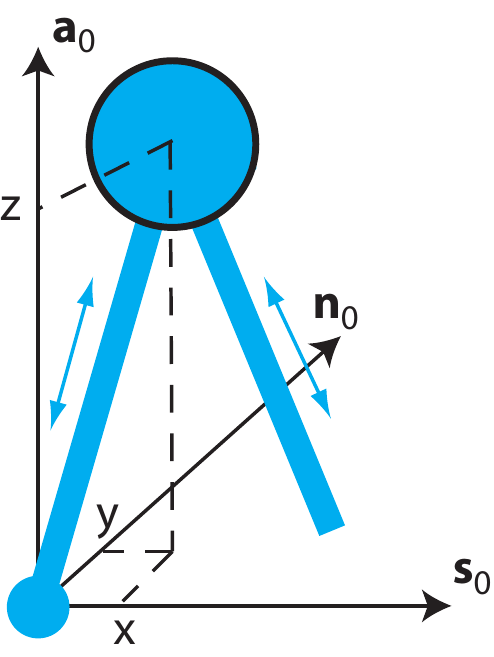}
        \caption{A simplified model of 3D biped robot. Each leg is
massless and has variable length. At the contact point, the stance leg
rotates passively around axes $\mathbf{s}_0$ and $\mathbf{n}_0$, the rotation around
axis $\mathbf{a}_0$ is not considered since this rotation is inhibited by
friction in normal biped locomotion. The swing leg has a fully actuated
spherical joint with respect to the concentrated mass of the hip.}
        \label{fig:rdouble_pendulum}
\end{figure}

In recognition of this, the proposed methodology is first presented
on a two-legged inverted pendulum, which consists of two
telescoping massless legs and a concentrated mass at their intersection, called the
hips;  see Figure
\ref{fig:rdouble_pendulum}. The stance leg is free to rotate about axes $\mathbf{s}_0$ and $\mathbf{n}_0$ at the
ground contact and the length of each leg can be modified through
actuation, allowing a desired vertical motion of the pendulum to be obtained.
 We also assume that the actuation of the
swing leg will allow a controlled displacement of the swing leg end
via hip actuators and control of the leg length.

The configuration of the robot is defined via the position of the
concentrated mass $(x_M, y_M, z_M)$ with respect to a reference frame
attached to the stance foot and the position of the swing leg tip, denoted by $
(x_s, y_s, z_s)$. Angular momenta along axes $\mathbf{s}_0$ and $\mathbf{n}_0$ are
denoted by $\sigma_x$ and $\sigma_y$. In order to explore
simultaneously the existence and stability of periodic orbits as a function of step length and width, a dimensionless dynamic model
of the pendulum will be used \cite{Chevallereau}. The normalizing
scaling factors applied along axes $\mathbf{s}_0$ and $\mathbf{n}_0$ depend on
desired step length $S$, desired step width $D$, and the mass $m$ of
the robot. Thus, a new set of variables is defined: $(X,Y,z_M,X_s, Y_s,
z_s,\sigma_X,\sigma_Y)=(\frac{x}{S},\frac{y}{D},z_M,\frac{x_s}{S},
\frac{y_s}{D}, z_s,\frac{\sigma_x}{mD},\frac{\sigma_y}{mS})$.

{\color{black}The variables
$(z_M, X_s, Y_s, z_s)$ will be the controlled variables $\aq$, whose evolution
will be determined by virtual constraints as a function of $\uq=(X,Y)$.}

The height of the CoM and its vertical velocity are
then expressed as functions of its horizontal position and velocity:
\begin{equation}\label{eq:z_pend}
\begin{aligned}
z_M^d&=f(X,Y)\\
\dot{z}_M^d&=\frac{\partial f(X,Y)}{\partial X}\dot{X}+\frac{\partial
f(X,Y)}{\partial Y}\dot{Y}
\end{aligned}
\end{equation}
Since the legs are massless, the
position of the swing leg tip and its time derivative do not affect
the dynamic model and thus have no effect on the evolution of the biped during single
support.
{\color{black} The moment balance equation of the pendulum around the rotation axis $\mathbf{s}_0$ and $\mathbf{n}_0$ directly yields the equation of the zero dynamics,
\begin{equation}\left\lbrace
\begin{aligned}
\dot{\sigma}_x&= -mgy_M,\\
\dot{\sigma}_y&=mgx_M.
\end{aligned}\right.
\end{equation}}
For this simple model, the angular momentum is simply
\begin{equation}\left\lbrace
\begin{aligned}
\sigma_x&= m \dot{z}_M y_M - m z_M \dot y_M,\\
\sigma_y&= - m \dot{z}_M x_M + m z_M \dot x_M,
\end{aligned}\right.
\end{equation}

{\color{black}Using the normalized coordinates, the zero dynamics model \eqref{eqn:CC:modeling:model_zeroAlternative} with  $z_2= \sigma_f=[\sigma_X,\sigma_Y]^{\top}$ can be written as
\begin{equation}\label{eq:Dyn_pend_cl}\left\lbrace
\begin{aligned}
\duq&=M_{XY}
^{-1}
\sigma_f, \\
\dot{\sigma}_X&= -gY,\\
\dot{\sigma}_Y&=gX,
\end{aligned}\right.
\end{equation}
where
$$M_{XY}=\begin{bmatrix}
\frac{\partial f(X,Y)}{\partial X}Y & \frac{\partial f(X,Y)}
{\partial Y}Y-f(X,Y)\\
-\frac{\partial f(X,Y)}{\partial X}X+f(X,Y)
& -\frac{\partial f(X,Y)}{\partial Y}X
\end{bmatrix}.$$}
We assume that once the tip of the swing leg hits the ground, that is, when $z \in \poincaresection \cap \zdmanifold$, the legs immediately swap their roles; thus, we assume that the
double support phase is instantaneous.

The state right before (after) the swapping of roles of the legs is denoted by
the exponent $^-$ ($^+$). During the swapping of the support leg,
the configuration of the robot is fixed, but the reference frame is
changed since it is attached to the new stance leg tip. From the
geometry, for normalized variables we have:
\begin{equation}\label{eq:impact_pos}\left\lbrace
\begin{aligned}
X^+ &= X^- - X_s^-,\\
Y^+ &= -Y^- + Y_s^-,
\end{aligned}\right.
\end{equation}
where the change of sign in the second equation corresponds to the
change of direction of the axis $\mathbf{n}_0$. Since the velocity of the CoM
of  a biped with massless legs is conserved in the transition, we
also have
\begin{equation}\label{eq:impact_vit}\left\lbrace
\begin{aligned}
\dot X^+ &= \dot X^-,\\
\dot Y^+ &= -\dot Y^-.
\end{aligned}\right.
\end{equation}

The design of the switching manifold $\mathbb{S}$ defined via $z_s^d(X,Y)$, and the transition condition \eqref{eq:impact_pos} affected by $X^-$, $Y^-$ defined via $X_s^d(X,Y)$ and $Y_s^d(X,Y)$, are discussed in the next section.

\subsection{Virtual constraints for the swing leg}
\label{sec:CompVirtCons}

{\color{black} Assume that a desired periodic motion is characterized by an initial free configuration for the single support phase $X_0, Y_0$ and a final free configuration $X_f, Y_f$. Due to the normalization of the variables with respect to the gait length and width, and the desired symmetry between left and right support, we have $X_f-X_0=1$ and $Y_0+Y_f=1$. Our initial objective is to define a virtual constraint for the height of the swing foot, $z_s^d$, as a function of $X$ and $Y$. This virtual constraint takes the value $0$ for $X=X_0, Y=Y_0$ and $X=X_f, Y=Y_f$, and is positive in between during the single support phase.}

\begin{figure}
\centering
        \subfigure[The mass motion starts on the switching manifold on
one stance leg]{\includegraphics[width =0.24\textwidth]{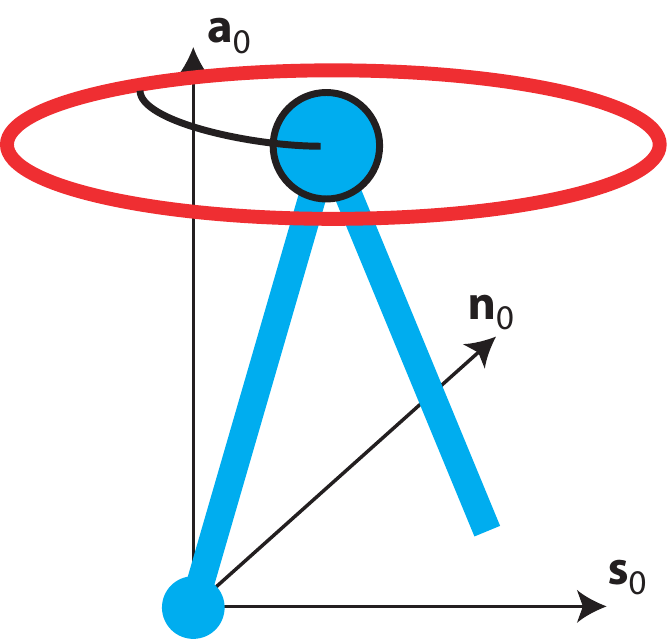}}
        \subfigure[When the mass crosses the switching manifold,
the support leg is changed]{\includegraphics[width =0.24\textwidth]
{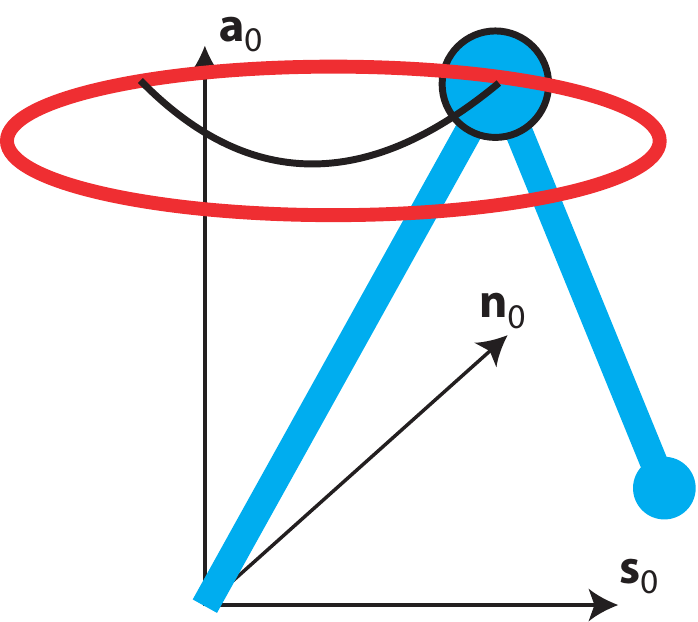}}
        \caption{The switching manifold (red ellipse)
defines the  position of the CoM where the swing leg hits the
ground, with an instantaneous change of support leg.}
        \label{fig:switchmanifold}
\end{figure}

{\color{black}Inspired by \cite{Razavi2014}, we propose 
  \begin{equation}\label{eq:z_gene}
  z_s^d= Z_{se}(X,Y) = \nu_z S_a(X,Y),
  \end{equation}
where $\nu_z$ is a negative parameter that adjusts the height of the swing foot during a step and $S_a(X,Y)$ is
\begin{equation}\label{eq:switch}
S_a(X,Y)=(X-X_a)^2+CY^2-((X_0-X_a)^2+CY_0^2)
\end{equation}
with
\begin{equation}
X_a =\frac{(X_f+X_0)+C (Y_f-Y_0)}{2}.\label{eq:xa}
\end{equation}
With this choice the ellipse-shaped switching manifold is
defined by
\begin{equation}\label{eq:switch2}
\mathbb{S}=\left\lbrace (X,Y)|S_a(X,Y)=0\right\rbrace,
\end{equation}
and is illustrated in Figure \ref{fig:switchmanifold}.}
\vspace{.1cm}

{\color{black} \textbf{Remark 2:} The shape of the switching manifold is somewhat arbitrary. We will see in the stability analysis done in the following section that only the tangent of the switching manifold at $X_f, Y_f$ will be used. The orientation of this tangent is characterized by the parameter $C$, and its effect will be studied in detail.}
\\
{\color{black} The virtual constraint imposed for the height of the swing foot defines the switching manifold and is crucial for the stability of a gait.
The virtual constraint associated with the horizontal position of the swing foot will allow us to impose where the swing foot is placed on the ground.}

{\color{black}In the case studied, the horizontal position of the swing foot is controlled to obtain an
($X_0, Y_0)$-invariant gait; that is, the starting position of the CoM with respect to the stance foot will be $(X_0, Y_0)$ independent of any error in the previous step. Based on \eqref{eq:impact_pos}, the horizontal motion of the swing leg must be such that
when $S_a(X,Y)=0$ (i.e., when the leg touches the ground), the
desired position $X_s^-=-X_0 + X$, $Y_s^-=Y_0+Y$ is reached. }

{\color{black}To achieve this objective, at the end of the
motion of the swing foot, the virtual constraints are defined by:
\begin{equation}
\label{eq:swing_leg_end}
   \begin{aligned}
   X_s^d = X_{se}(X,Y) = \left( 1- \nu_X S_a(X,Y)\right) \left(-X_0+X\right),\\
  Y_s^d = Y_{se}(X,Y) = \left( 1- \nu_Y S_a(X,Y) \right) \left(Y_0+Y\right).
   \end{aligned}
\end{equation}
With these constraints, for any values of $\nu_X$ and $\nu_Y$, the
desired initial position of the CoM at the beginning of the next step, that is $X_0$ and $Y_0$,  will be obtained. Moreover, in order to increase the robustness with respect to variations of the ground height, the
parameters $\nu_X$ and $\nu_Y$ are chosen such that in a periodic gait at the end of the step the horizontal velocity of the swing leg tip is zero.}

{\color{black} We note that \eqref{eq:swing_leg_end} expresses the desired
horizontal position of the swing foot at the end of the step. At the beginning of the step, for $X=X_0, Y=Y_0$, equation \eqref{eq:swing_leg_end} obviously does not provide the initial position of the swing leg tip, $X_S^+=-1$, $Y_S^+=1$ for the periodic gait. To solve this issue, the
virtual constraints concerning the motion of the swing leg are in
fact decomposed into two parts: one concerning near the end of the
step, more specifically, for
$X>X^l=0.4+\frac{X_0+X_f}{2}$ to the end of the step and one concerning the beginning of the step for
which a five order polynomial function of $X$ is added to the
virtual constraints of the second  part. Therefore, the motion along
the axis $\mathbf{s}_0$ is imposed by the following equations:
\begin{equation}
\begin{array}{ccc}
   \label{eq:cons_swing_leg_end}
   X_s^d &=& X_{se}(X,Y) + P_x(X) ~\mathrm{ for }~ X < X^l, \\
  X_s^d &=& X_{se}(X,Y) ~\quad\quad\quad\quad\mathrm{ for } ~X \ge X^l.\\
     Y_s^d &=& Y_{se}(X,Y) + P_y(X) ~\mathrm{ for }~ X < X^l, \\
  Y_s^d &=& Y_{se}(X,Y) ~\quad\quad\quad\quad\mathrm{ for } ~X \ge X^l.
\end{array}
\end{equation}
The coefficients of the polynomial $P_x$ are such that at $X=X^l$,
continuity in positions, velocities and accelerations are ensured
(that is, $P_x(X^l)=0$, $\frac{\partial P_x(X^l)}{\partial X}=0$, $
\frac{\partial^2 P_x(X^l)}{\partial X^2}=0$, $P_y(X^l)=0$, $\frac{\partial P_y(X^l)}{\partial X}=0$, $
\frac{\partial^2 P_y(X^l)}{\partial X^2}=0$.), and at the beginning
of the step, the desired position and velocity are obtained, i.e., $X_s^-=-1$, $Y_s^-=1$, $\dot X_s^-=0$, $ \dot Y_s^-=0$. An illustration of the use of a polynomial function to ensure continuity of the virtual constraint is shown in Figure \ref{fig:zcor}. }

{\color{black} In fact, for the height of the swing foot, \eqref{eq:z_gene} is used also only at the end of the step. At the beginning of the step, a polynomial function of $X$ is also added. This modification allows us to take into account that since the leg was previously in support, its initial velocity is zero. A fourth order polynomial is used to ensure the continuity in position and velocity at the middle of the step:
\begin{equation}
\begin{array}{ccc}
   z_s^d &=& Z_{se}(X,Y) + P_z(X) ~\mathrm{ for }~ X < \dfrac{X_0+X_f}{2}, \\\\
  z_s^d &=& Z_{se}(X,Y) ~\quad\quad\quad\quad\mathrm{ for } ~X \ge \dfrac{X_0+X_f}{2}.
\end{array}
\end{equation}
This slight modification does not change the switching surface.}
\\
\vspace{.1cm}

\textbf{Remark 3}: The selection of the parameters  $X_f$, $Y_f$, $\dot X_f$, $\dot Y_f$, $\nu_X$, and $\nu_Y$ depends on our expected duration of the periodic step. These parameters are found by solving a boundary value problem describing the periodicity of the gait. A detailed illustration of the swing leg trajectory is given in Section \ref{sec:huma_cyc}.

\subsection{Gait with constant height and self-synchronization}\label{sec:constant}

An inverted pendulum with a constant height
$z=z_0$ is known as a 3D linear inverted pendulum (3D LIP)
\cite{Kajita2001}.
Since with the condition $z = z_0$ the motions in the sagittal and frontal planes are decoupled,
from \eqref{eq:Dyn_pend_cl}, the expression of the 3D LIP hybrid zero dynamics reduces to:
\begin{equation}
\label{eq:3dlip_continuous_dynamics}
\left\lbrace
\begin{aligned}
\dot{X}&= \frac{\sigma_Y}{z_0},\\
\dot{Y}&= -\frac{\sigma_X}{z_0},\\
\dot{\sigma}_X&= -gY,\\
\dot{\sigma}_Y&=gX,
\end{aligned}\right.
\end{equation} which can be analytically solved.

\begin{figure}[h]
\begin{center}
\includegraphics[scale=.4]{\images 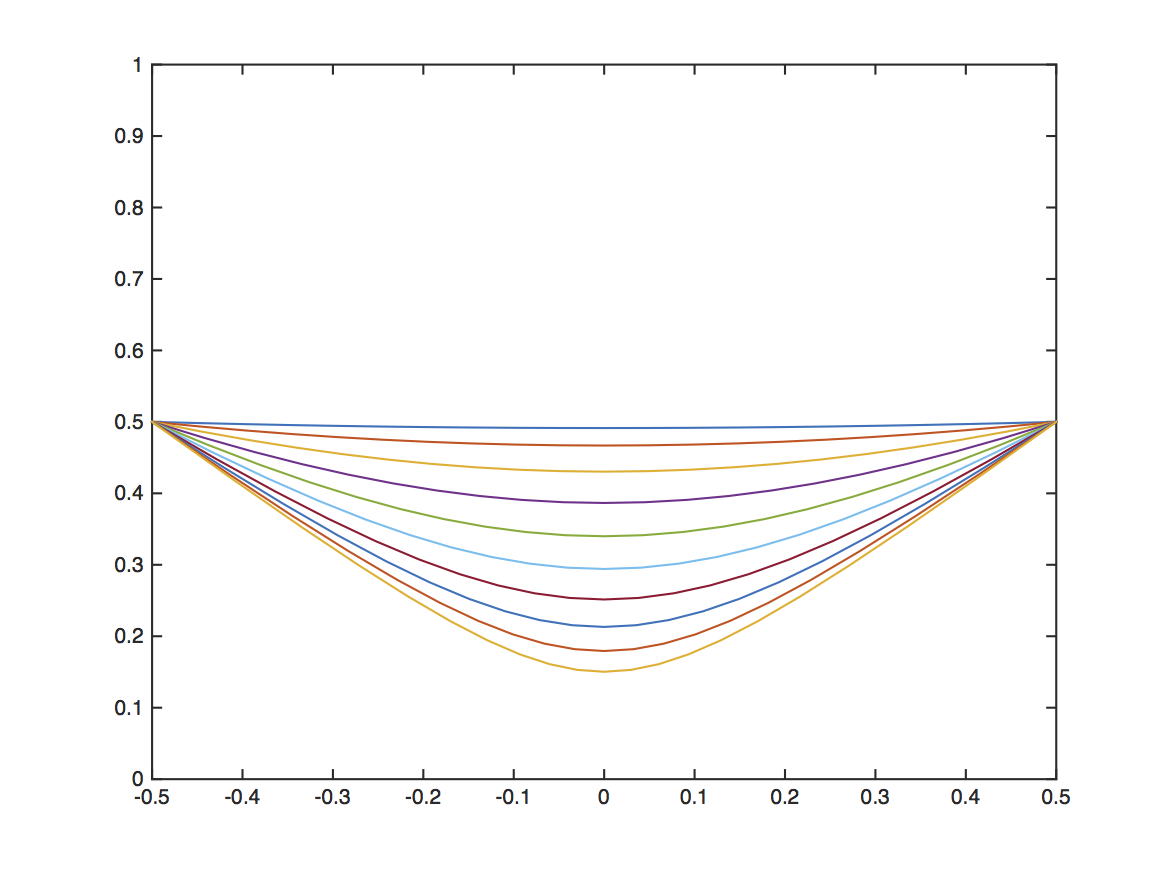}
\end{center}
\caption{Periodic motions in normalized variables for  several values of $T$, orientation of the final velocity is defined by $0 <-\frac{\dot Y_0}{\dot X_0}< 1$.}
\label{cycle_fig}
\end{figure}

The notion of self-synchronization for the 3D LIP model, which was defined and characterized in \cite{Razavi2014}, is briefly recalled.
{\color{black} For a 3D LIP model, it is well known that the orbital energies:
\begin{equation} \label{eq:cons_E}
\begin{aligned}
\mathcal{E}(X)=\dot{X}^2-\frac{g}{z_0}X^2\\
\mathcal{E}(Y)=\dot{Y}^2-\frac{g}{z_0}Y^2
\end{aligned}
\end{equation}
are conserved quantities during a step \cite{Kajita2001}.
Moreover, the quantity
\begin{equation}\label{eq:cons_L}
\mathcal{L}(X,Y)=\dot{X}\dot{Y}-\frac{g}{z_0}XY
\end{equation}
is also conserved and is called the synchronization measure
\cite{Razavi2014}. If this quantity is zero, the motion between the
sagittal and frontal planes is synchronized (i.e., $\dot{Y}=0$ when
$X=0$ \cite{Razavi2014}). Any periodic motion of the LIP, with symmetric motion for left and right support, is characterized by
\begin{equation}
X_0=-\frac{1}{2},~~X_f=Y_0=Y_f=\frac{1}{2}.\label{eq:X0XfY0Yf}
\end{equation}
In normalized variables, the periodic motion for different values of step durations $T$ are presented in Figure \ref{cycle_fig}.}

{\color{black} In \cite{Chevallereau}, it is shown that for a periodic motion of the LIP, self-stabilization  is impossible. In this section, our objective is to show the influence of the virtual constraints chosen for the swing leg, and especially the influence of the switching manifold on the self-synchronization of the walking gait.
Indeed, since synchronization between the sagittal and frontal motions implies coupling between these two motions and since these motions are decoupled during the single support phase, it is essential to introduce a coupling at the transition. }

{\color{black} For the 3D LIP model, due to the specific values of $X_0,X_f,Y_0, Y_f$, the switching configuration manifold corresponding to the virtual constraints defined in \eqref{eq:switch2} becomes:
\begin{equation}\label{eq:switchLIP}
\mathbb{S}=\left\lbrace (X,Y)|X^2+CY^2-(X_0^2+CY_0^2)=0\right\rbrace
\end{equation}
because $X_a=0$ using \eqref{eq:xa} with \eqref{eq:X0XfY0Yf}.}
%

The virtual constraints for the swing legs are also designed such that $(X^+, Y^+) = (X_0,
Y_0)$ at the beginning of the ensuing step, so the 3D LIP is said to be $(X_0,Y_0)$-invariant. In scaled coordinates, we limit our study to the case $X_0=-\frac{1}{2}$ and $Y_0=\frac{1}{2}$.

For an $(X_0,Y_0)$-invariant 3D LIP, if the switching manifold is
defined by \eqref{eq:switchLIP}, reference \cite{Razavi2016} shows
that the Jacobian of the Poincar\'e return map at the fixed point,
 expressed in the coordinate system $(X_k^-, \mathcal{L}_k, \mathcal{K}_k^-)$, where $\mathcal{L}_k$ is the synchronization measure at the end of step $k$ and $\mathcal{K}_k^-$ is the kinetic energy at the end of the step $k$, is in the form
\begin{equation}\label{Jacobian} \mathbf{J} = \begin{bmatrix}
0 & * & 0\\
0 & -\lambda_{\mathcal{L}} & 0\\ 0&*&1
\end{bmatrix},
\end{equation}
where $*$ represents non-zero terms.
Three eigenvalues are identified in this Jacobian. One is zero due to the fact that the position of the CoM at the end of
one step does not affect the dynamics of the following step for an $
(X_0,Y_0)$-invariant 3D LIP. One is a unit eigenvalue which
corresponds to the kinetic energy of the pendulum, which indicates that
a variation of the kinetic energy will be conserved, and the system
will not converge to its original fixed point.

{\color{black} We will propose in this paper a new expression for $\lambda_{\mathcal{L}}$, and we will give an original condition on the choice of the switching manifold to induce synchronization and a physical interpretation of this condition.}

{\color{black} Suppose that the state of the robot is slightly perturbed around the initial periodic state: $[-1/2, 1/2, \dot X_0, \dot Y_0]$. After $i$ steps the initial state of the robot is denoted by $[-1/2, 1/2, \dot X_i, \dot Y_i]$, and its synchronization measure is denoted $L_i$. After the following step the state will be $[-1/2, 1/2, \dot X_{i+1}, \dot Y_{i+1}]$, and its synchronization measure is denoted by $L_{i+1}$. The calculation of ${\frac{L_{i+1}}{L_i} }$ is developed in the appendix. The derivation is based on the conservation of orbital energies \eqref{eq:cons_E}, the synchronization measure  \eqref{eq:cons_L} during a single support phase, and the fact that the change of support occurs on the switching manifold \eqref{eq:switchLIP}. For small variation around the periodic orbit, the final error for each step satisfies:
\begin{equation}
\delta X^-_i = - C \delta Y^-_i.
\end{equation}}
{\color{black} Direct calculation gives:
\begin{equation}
\lambda_{\mathcal{L}} =\lim_{i \rightarrow \infty} {\frac{L_{i+1}}{L_i} }=  \frac{(\dot Y_0-\dot X_0)( C \dot Y_0 + \dot X_0)}{(\dot X_0+\dot Y_0)(-C \dot Y_0 + \dot X_0)}
\label{eq:conv_L}
\end{equation}}

{\color{black} Two factors appear in \eqref{eq:conv_L}: one varies as a function of $C$ and can be modified by the design of the switching manifold, while the other depends on the periodic gait velocity, i.e on the step duration of the gait.}


{\color{black} When $C \dot Y_0 + \dot X_0 = 0$, the synchronization occurs in one step. This case corresponds to a switching line co-linear with the initial velocity of the periodic motion since in this case we have:
\begin{equation} \label{eq:C_opt}
C= -\frac{\delta X^-_i}{\delta Y^-_i}=-\frac{\dot X_0}{\dot Y_0}
\end{equation}
Thus, the cross product between the final error in position $(\delta X^-_i, \delta Y^-_i)$ and initial periodic velocity $(\dot X_0, \dot Y_0 )$ is zero.}

{\color{black} The values of $C$ that ensure synchronization of the motion must be such that:
 \begin{equation}
\left | \lambda_{\mathcal{L}} \right | = \left |\frac{(\dot Y_0- \dot X_0)( C \dot Y_0 + \dot X_0)}{(\dot X_0+\dot Y_0)( -C \dot Y_0 + \dot X_0)}\right |<1 \label{eq:normlambda}
\end{equation}}
{\color{black} For a  periodic velocity corresponding to Figure \ref{cycle_fig}, namely, when $\dot X_0 >0$, $\dot Y_0<0$, $\dot X_0 > -\dot Y_0$, it can be shown that $C$ has to satisfy:
\begin{equation}\label{eq:lambda_L1}
1 < C < \left (\frac{\dot X_0}{\dot Y_0} \right)^2.
\end{equation}}

\begin{figure}
        \centering
        \subfigure[]{\includegraphics[width=0.24\textwidth]{\images
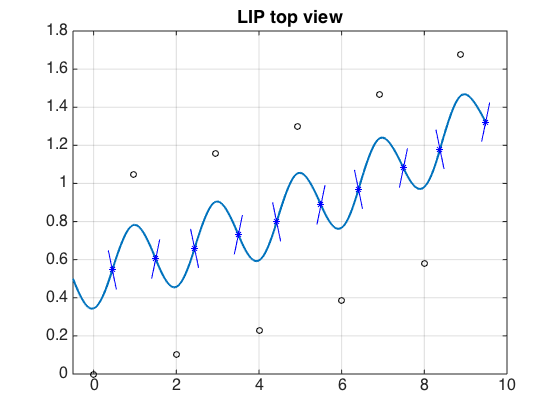}}
        \subfigure[]{\includegraphics[width=0.24\textwidth]{\images
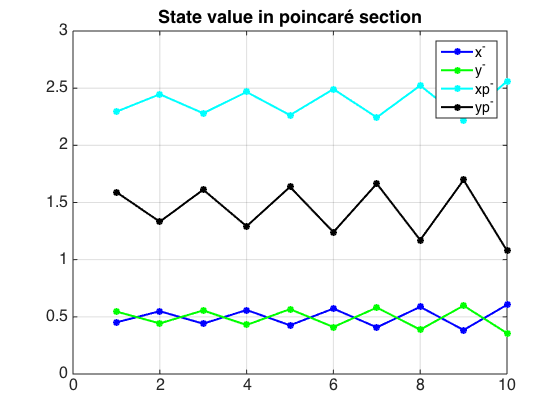}}
\subfigure[]{\includegraphics[width=0.24\textwidth]{\images
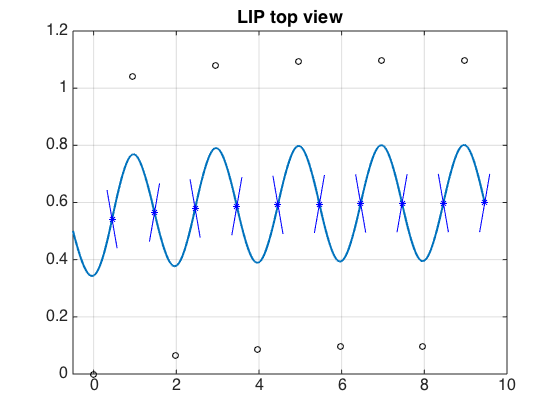}}
\subfigure[]{\includegraphics[width=0.24\textwidth]{\images
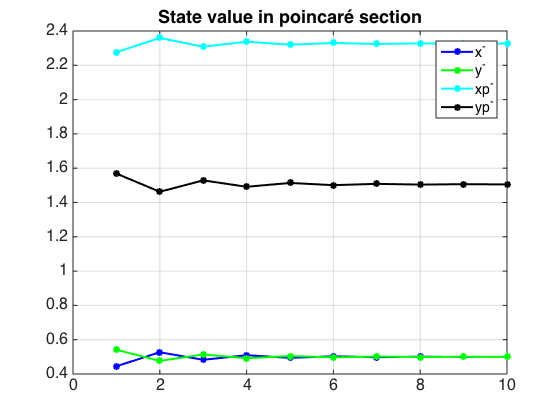}}
\subfigure[]{\includegraphics[width=0.24\textwidth]{\images
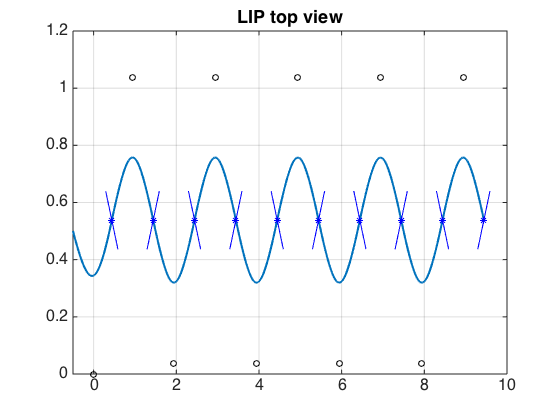}}
\subfigure[]{\includegraphics[width=0.24\textwidth]{\images 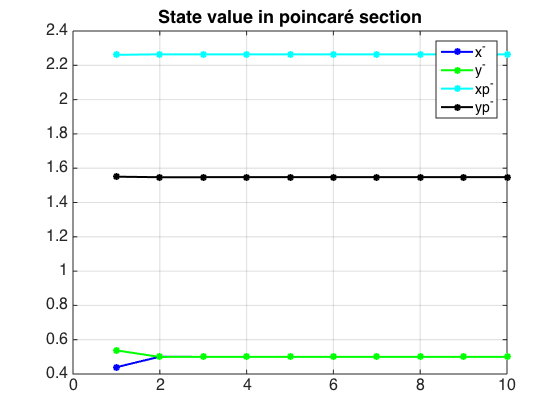}}
        \caption{{\color{black} Starting from the same initial state $X_0=-0.5$, $Y_0=0.5$, $\dot X_0=2.3147~s^{-1}$, $\dot Y_0=-1.5136~s^{-1}$, the simulation of 10 steps shows  several behaviors depending on the switching manifold defined by $C$. From top to bottom, we have $C=0.95$; $C=1.2$, $C=1.45$. The left-hand figures show the position of the stance foot, the evolution of the CoM, and a part of the switching manifold  are shown. On the right-hand figures, the state of the HZD is shown in the Poincar\'e section, just before the change of support: divergence or convergence is clearly illustrated.}}
        \label{fig:cond_synchro}
 \end{figure}

{\color{black} In Figure \ref{fig:cond_synchro}, for the same initial state, the behavior of the LIP is shown for several values of C. Three cases are illustrated for C satisfying \eqref{eq:lambda_L1} or not.  For $C=0.95$, since condition \eqref{eq:lambda_L1} is not satisfied, a periodic motion is not obtained, and the direction of walking is not along the axis $s_0$. For the two other cases tested, the condition \eqref{eq:lambda_L1} is satisfied, and convergence to a periodic motion is observed, but in the two cases, the periodic gait velocities differ. In the case $C=1.45$, since the value of $C$ is close to \eqref{eq:C_opt}, the synchronization is faster than in the case $C=1.2$. It should be noted that the values of $\dot X_0, \dot Y_0$ correspond to the periodic motion. In the case $C=1.2$, the LIP motion converges to a motion such that $\dot X_0=2.3265~s^{-1}$, $\dot Y_0=-1.5059~s^{-1}$, thus, $-\frac{\dot X_0}{\dot Y_0}=1.5449$. In the case $C=1.45$, the LIP motion converges to a motion such that $\dot X_0=2.2639~s^{-1}$, $\dot Y_0=-1.5476~s^{-1}$, thus, $-\frac{\dot X_0}{\dot Y_0}=1.4628$.}

\begin{figure}
\centering
        \includegraphics[width=0.49\textwidth]{\images
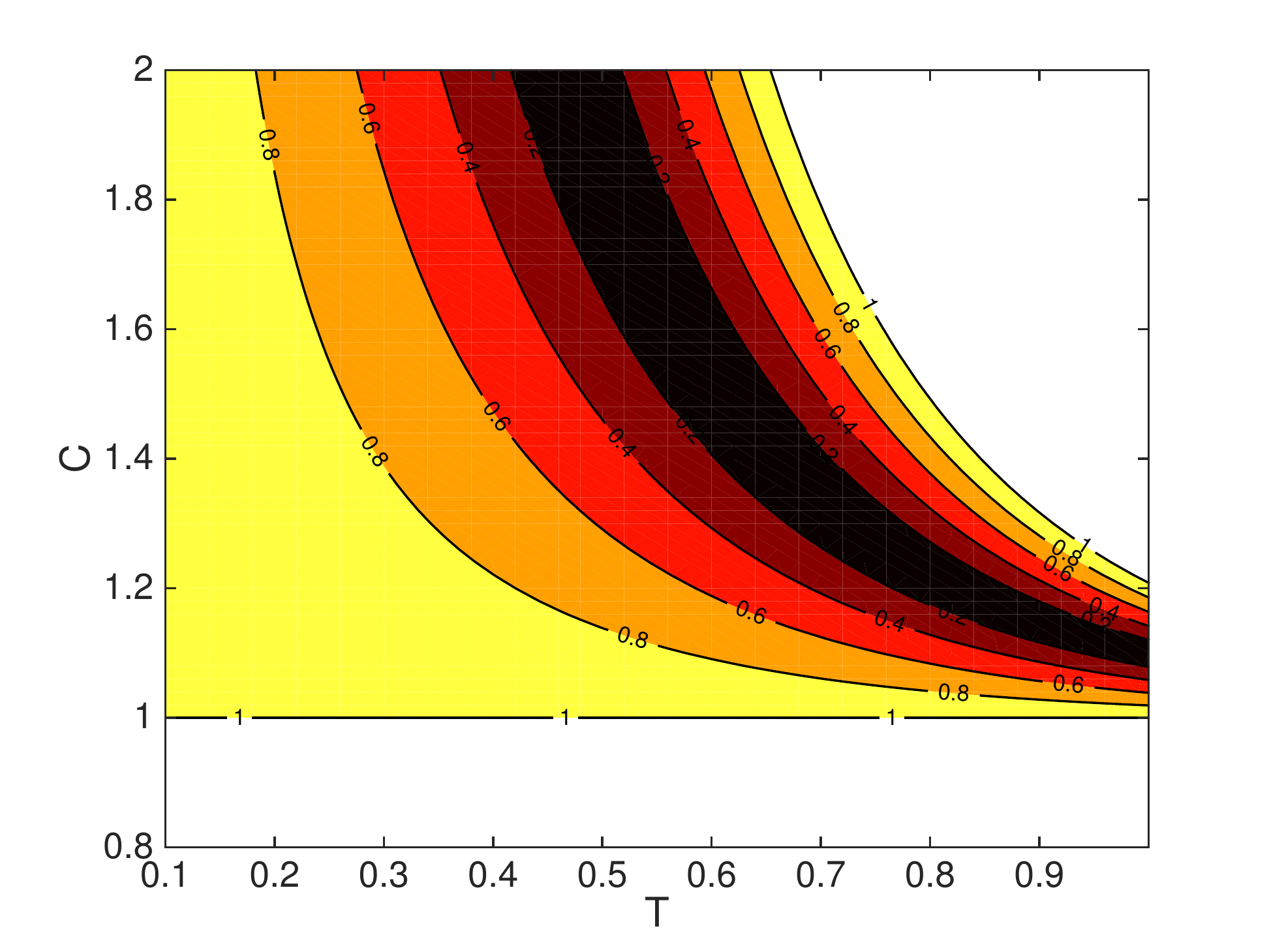}
        \caption[Level contours of $\lambda_{\mathcal{L}}$  expressed
as function of $C$ and step duration $T$ for the LIP model]{$
\lambda_{\mathcal{L}}$ level contours expressed as function of $C$
and step duration $T$ for the LIP model and $z_0=0.7$~m. }
        \label{fig:A_0}
\end{figure}

In Figure \ref{fig:A_0} the level contours of $|
\lambda_{\mathcal{L}}|$ are drawn as a function of the parameter $C$
and the step duration $T$.
The condition $|\lambda_{\mathcal{L}}|<1$ ensures convergence toward
a synchronized motion, thus is a condition of self-synchronization. However, the walking velocity of the gait is not
controlled in the sense that a perturbation of such a gait will
result in the convergence of the gait back to a periodic motion,
but with a different gait velocity {\color{black} as illustrated in Figure \ref{fig:cond_synchro}}. In the next section, we will see how judiciously adding oscillations of the CoM can stabilize the walking velocity.

\subsection{VLIP model and self-stabilization}
\label{sec:VLIP-stab}

%
%
From study of planar walking robots, it is known that vertical
oscillations of the CoM can asymptotically stabilize a periodic
walking gait \cite{Westervelt2007a}. Thus, here, in order to stabilize the walking velocity in an $(X_0,Y_0)$-invariant gait, oscillations of the CoM will
be introduced. These oscillations of the CoM are obtained by the following virtual constraint:
\begin{equation}\label{eq:constraint}
z^d(X,Y)=z_0-aS_a(X,Y),
\end{equation}
with $S_a(X,Y)$ defined in \eqref{eq:switch} associated to the
ellipse-shaped switching manifold defined in \eqref{eq:switch2}

The expression of $S_a$ ensures that at the beginning and end of the step $z^d=z_0$ since $S_a(X_0,Y_0) = S_a(X_f,Y_f) =0$. Note that the case $a=0$ with $X_a = 0$
corresponds to the 3D LIP example described in Section
\ref{sec:constant}.

\begin{figure}
        \centering
           \psfrag{X}{$X$}
           \psfrag{Y}{$Y$}
              \psfrag{z}{$z$}
        \includegraphics[width=0.49\textwidth]{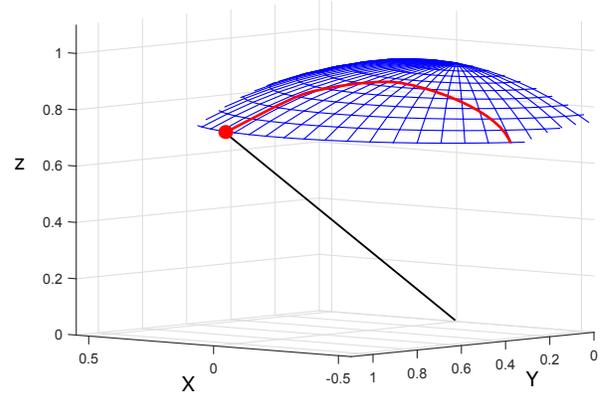}
        \caption{Surface defined by the virtual constraint for the
vertical motion the mass motion $z^{d}(X,Y)$. The red curve gives an
example of the mass evolution during a single support phase.}
        \label{fig:zsurf}
\end{figure}

During the single-support phase, the horizontal position of the mass
is located inside the ellipse; thus, $S_a(X,Y)$ is negative and
increasing when approaching the switching manifold ellipse. Choosing
$a>0$ will ensure a negative vertical velocity of the mass at the
transition (see Figure \ref{fig:zsurf}).
This negative vertical velocity of the CoM right before the change
of support implies that the angular momenta $\sigma_X$ and $
\sigma_Y$ decrease at the change of support. In order to obtain a
periodic motion, the angular momenta must therefore increase during the stance
phase, and thus it is necessary to slightly shift the relative position
of the support leg and the CoM \cite{Chevallereau2003}. The position
of the CoM at the beginning of the step is then written as $X_0=-
\frac{1}{2}+D_X$, $Y_0=\frac{1}{2}-D_Y$ and at the end of the step
$X_f=\frac{1}{2}+D_X$, $Y_f=\frac{1}{2}+D_Y$ with $D_X = \frac{X_0+ X_f}{2} \geq 0$ and
$D_Y=\frac{Y_f-Y_0}{2} \geq 0$.

\begin{figure}
        \centering
                    \psfrag{X}{$X$}
              \psfrag{z}{$z$}
                    \psfrag{0}{$0$}
        \includegraphics[width=0.30\textwidth]{\images
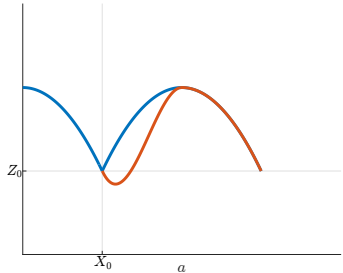}
        \caption{Planar illustration of the modification of the
virtual constraint at the beginning of each step to avoid
discontinuity in tracking error. At the beginning of the step, the
virtual constraint, in black, is continuous for the height of the CoM, but the velocity is not continuous. A modification of the
virtual constraint $z_{cor}(X)$ is added to satisfy the conditions
of continuity, the constraint effectively used is shown in red.}
        \label{fig:zcor}
\end{figure}

This choice of the switching
manifold only ensures the continuity of the vertical position of the
CoM. To ensure the continuity of the  vertical velocity of the CoM
as well, a third order polynomial function of $X$, denoted by
$z_{cor}(X)$, is added to expression (\ref{eq:constraint}) for
$z^d$. This function is null at the beginning of the step, where
$X=X^+$, because the continuity of the position is already ensured,
but its derivative should compensate the difference between the
vertical velocity at the end of the previous step and the one
corresponding to the reference $\dot z^{d}(X_0,Y_0)$ in
(\ref{eq:constraint}). This correction term only acts from the
beginning of the step where $X=X^+$ to the mid-step where $X=D_X$.
In the second half of the step, where $X \ge D_X$,  $z_{cor}(X) \equiv 0$, and smoothness in the mid-step is guaranteed by imposing
the conditions $z_{cor}(D_X)= \dot z_{cor}(D_X)=0$. This technique,
which has already been used in previous studies \cite{MoGri2009,
Chevallereau2008}, is illustrated in Figure \ref{fig:zcor}.

In the presence of vertical oscillations, periodic
motions will exist only for an appropriate set of virtual
constraints. In particular, the values of $D_X$ and $D_Y$ cannot be
chosen arbitrarily.  In practice, the values of $C$, $a$, and $T$, the
duration of the periodic step, are chosen, and $D_X$, $D_Y$, and the angular momentum at the end of the single support (or the final velocity of the CoM $(\dot X^+$, $\dot Y^+$) are
deduced by solving a boundary value problem describing the
periodicity of the evolution of the CoM. The variations of
$D_X$ and $D_Y$ as functions of $a$ for fixed values of $C$ and
$T$ are shown in Figure  \ref{fig:DsDd}.

\subsection{Stability of the gait obtained}

\label{sec:stability_res}
The study of the gait stability is undertaken through numerical
calculations. The evolution of the mass in single support is
integrated numerically from the dynamic model
\eqref{eq:Dyn_pend_cl} taking into account the conditions \eqref{eq:constraint} and
\eqref{eq:switch2} developed in Section \ref{sec:VLIP}.

\begin{figure}
\centering
\subfigure[\label{fig:E}]{\includegraphics[width = 0.24\textwidth]
{\images 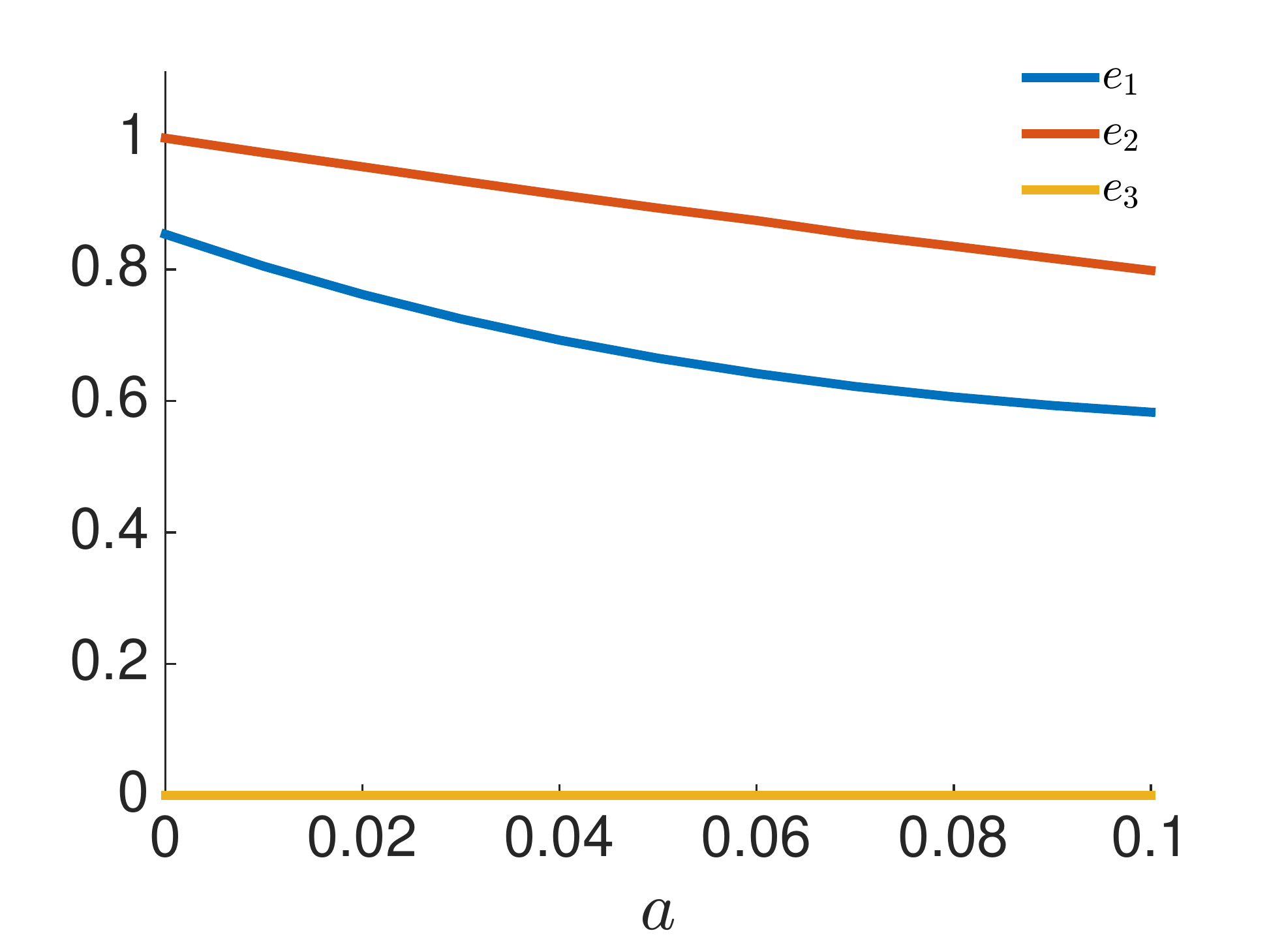}}
\subfigure[\label{fig:DsDd}]{\includegraphics[width =
0.24\textwidth]{\images 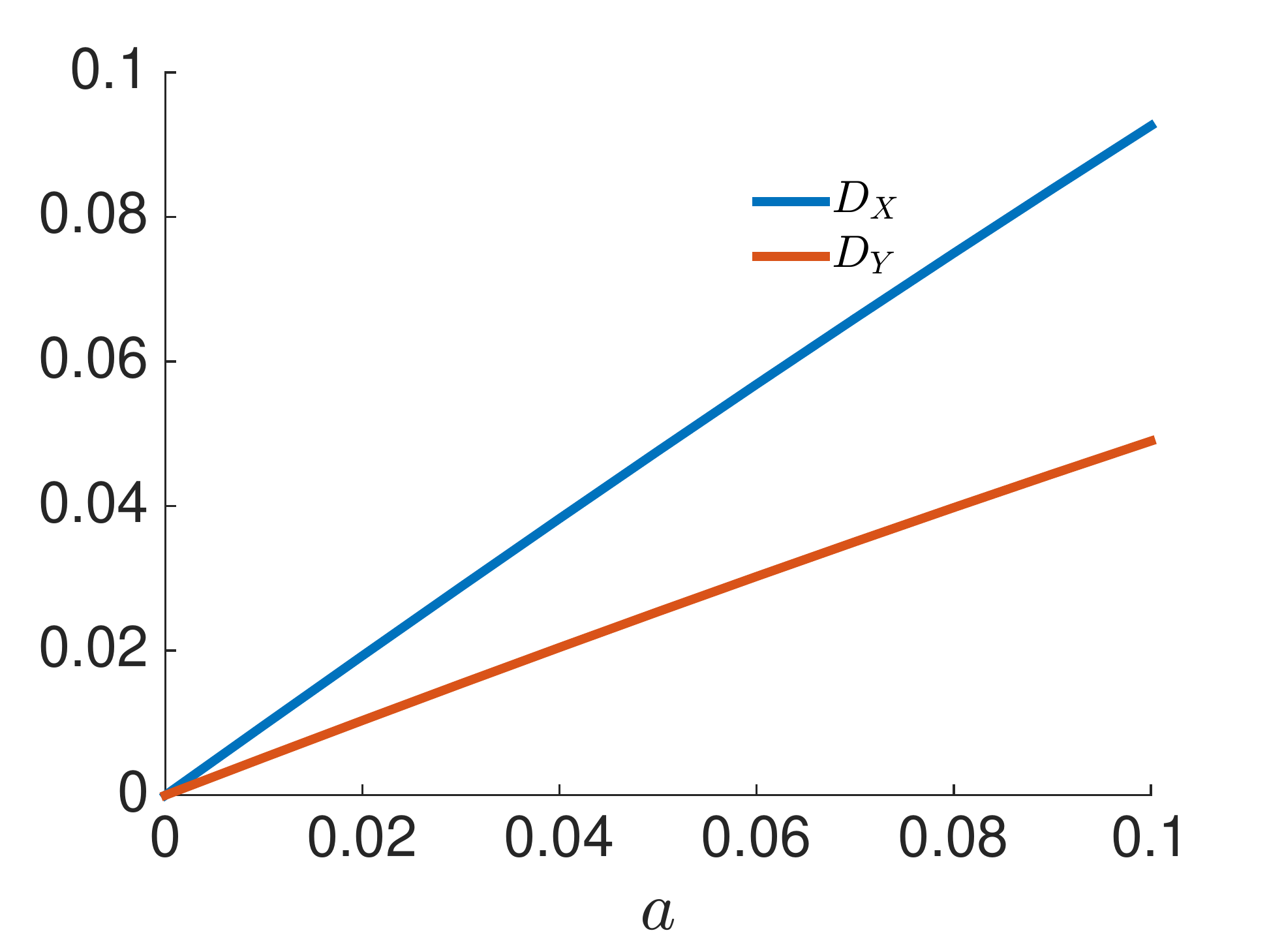}}
\caption{Influence of the amplitude of vertical oscillation
characterized by $a$ on the stability via the magnitude of the three
eigenvalues $e_1, e_2, e_3$ (a) and on the support foot position shift (b).
Illustrations are numerically obtained for the case $z_0=0.7$~m,
$C=1.1$ and $T=0.7$~s.}
\end{figure}

\begin{figure}
        \centering
              \psfrag{a}{$a$}
                    \psfrag{C}{$C$}
        \includegraphics[width=0.49\textwidth]{\images 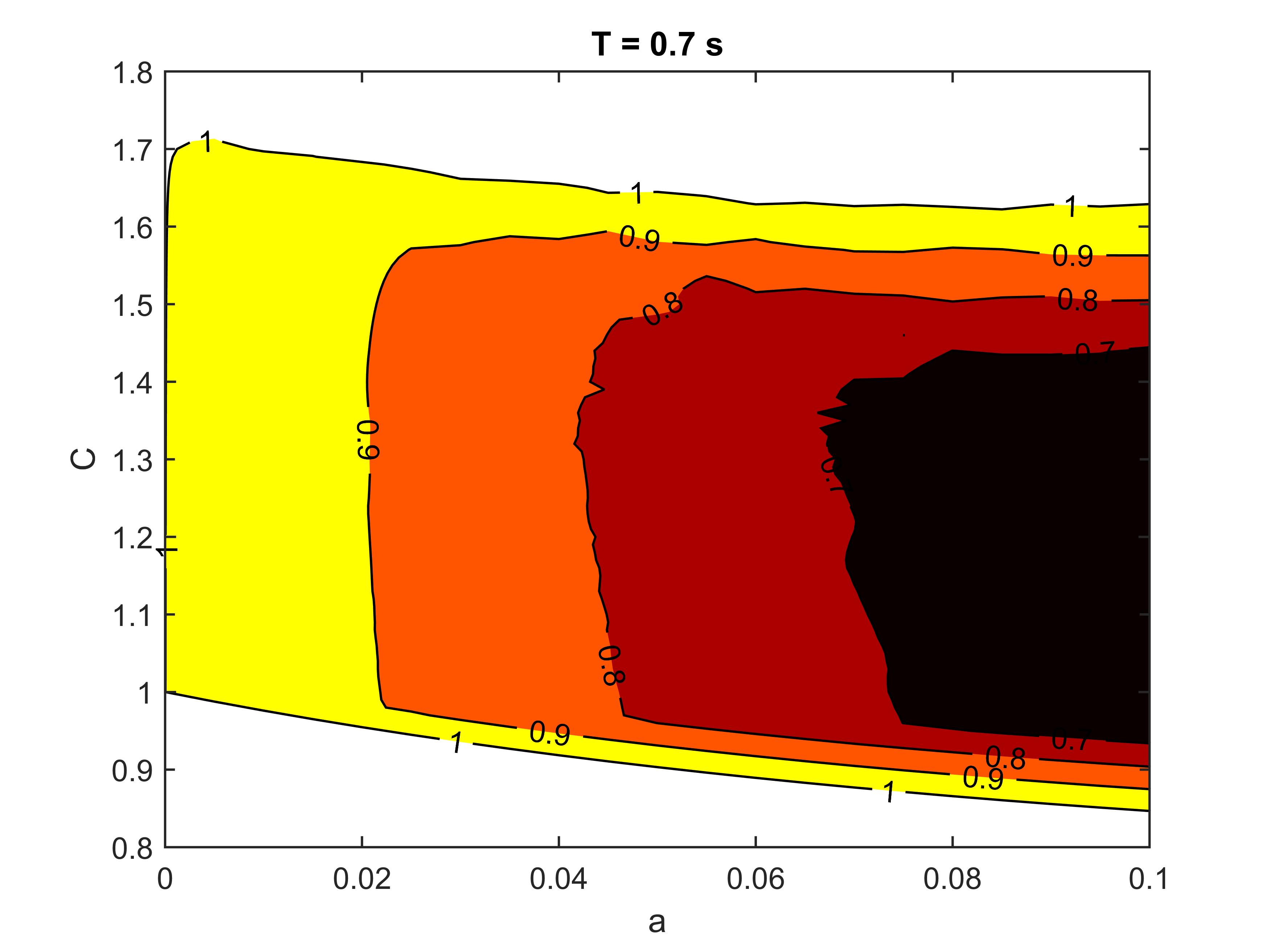}
        \caption{Stability criterion level contours for the
pendulum model as function of $C$ and amplitude parameter $a$ for
$T=0.7$~s and $z_0=0.7$~m.}
        \label{fig:Tfix}
\end{figure}

\begin{figure}
\centering
  \psfrag{T}{$T$}
\psfrag{C}{$C$}
        \includegraphics[width=0.49\textwidth]{\images 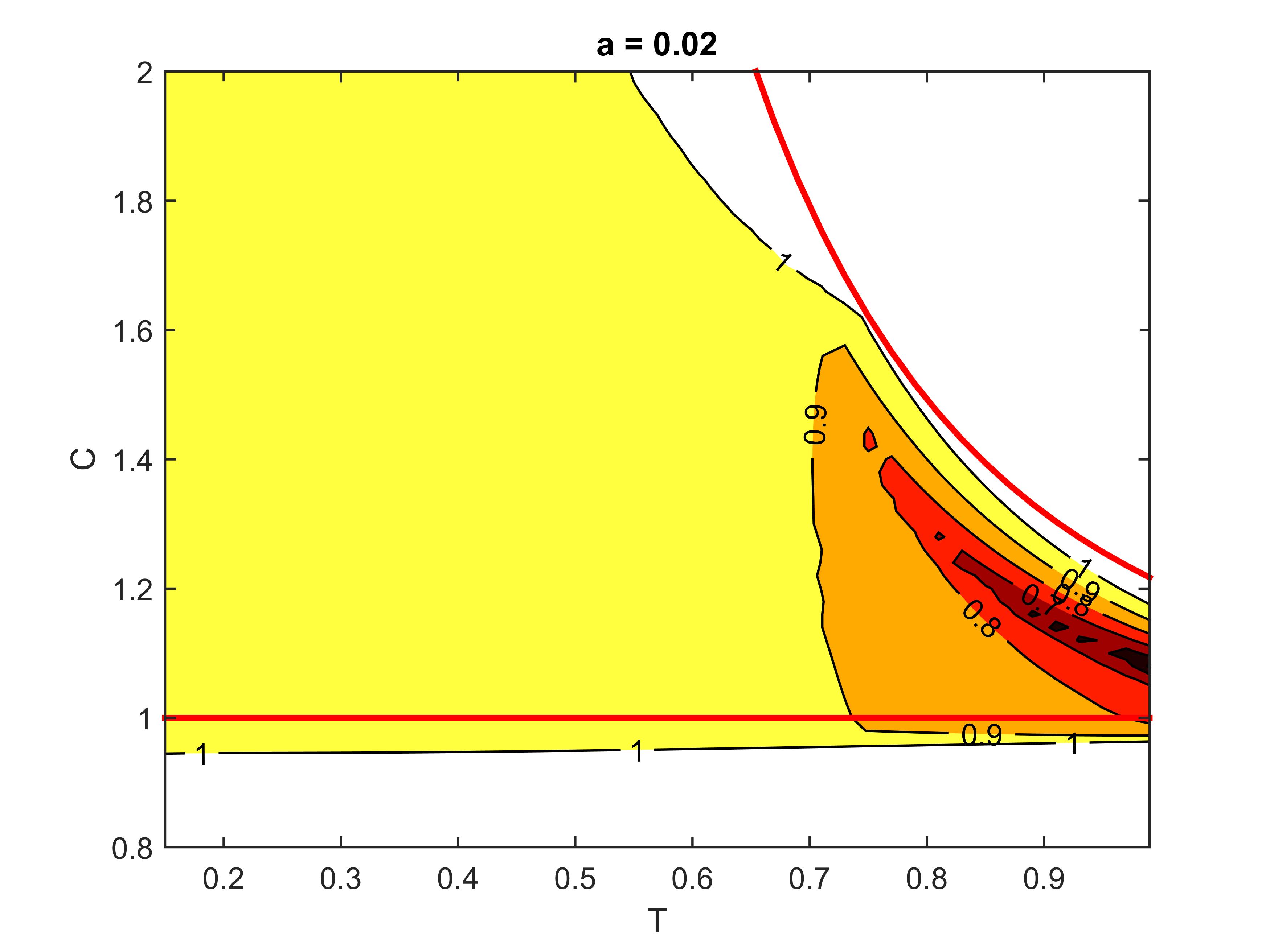}
        \caption{Stability criterion level contours for the
pendulum model as function of $C$ and step duration $T$ for
$a=0.02$~m and $z_0=0.7$~m. The red curves represent the boundaries
inside which $|\lambda_\mathcal{L}|<1$ (equation
(\ref{eq:lambda_L1})) for the LIP.}
        \label{fig:A_0_02}
\end{figure}

To study the effect of oscillations on stability, we first revisit
the 3D LIP case ($a=0$). The virtual constraints are completely
defined by the value of $z_0$ and $C$. We fix $z_0=0.7~m$ and will
study the effect of the parameter $C$. Many periodic orbits exist for $a=0$ corresponding to
various step durations $T$, however, for a given value of $T$, a unique
periodic gait exists. As explained in Section \ref{sec:constant}, three
eigenvalues are evaluated to examine the stability of the periodic
motion. In our numerical study, the Poincar\'e
return map is expressed using our state variables $\boldsymbol{\chi}_k$. Because the eigenvalues of the Jacobian of the Poincar\'e map are invariant under a change of coordinates, the results that we observed are identical to those of
 \eqref{Jacobian}: among the three eigenvalues, one is equal
to zero, one is equal to 1, and the last one is $\lambda_{\mathcal{L}}$. Selecting $C$ and step duration $T$ such
that $|\lambda_{\mathcal{L}}|<1$ according to \eqref{eq:lambda_L1} will allow us to observe the effect of
oscillations on the eigenvalues. Figure \ref{fig:E}
illustrates the evolution of the three eigenvalues $e_1, e_2$, and  $e_3$ for $C=1.1$ and $T=0.7$~s as the oscillation amplitude parameter $a$ increases. In our example, the magnitudes of the two non-null
eigenvalues decrease as the amplitude of the oscillations increases.
Thus, when $a>0$, the absolute values of all eigenvalues
are smaller than 1. Consequently, we observe that introducing
oscillations transforms the self-synchronization property of the walking
gait into self-stabilization. Figure \ref{fig:Tfix} shows the effect
of oscillations on stability; the stability criterion $\delta$ (defined as
the maximum magnitude of the eigenvalues) is shown via level contours for several values of the ellipse
parameter $C$ and a given step duration $T=0.7$~s. The value $\delta$ decreases when $a$ increases.  We observe that
when oscillations are introduced, the values of the ellipse
parameter $C$ for which stability exists are close to the ones where
self-synchronization is observed for the 3D LIP. Figure
\ref{fig:A_0_02}
shows the evolution of the stability criterion $\delta$ for $a=0.02$~m. We
can see that the stability effect observed in our first example
(Figure \ref{fig:E}) is extended to most of the values of the parameter
$C$ and step duration $T$.
To complete this analysis, Figure \ref{fig:DsDd}
shows the evolution of the parameters $D_X$ and $D_Y$, which
introduce asymmetry in the gait, obtained for a periodic motion as the
amplitude of the oscillations increases. As expected, the values of $D_X$ and $D_Y$ increase with
oscillations to compensate the loss of angular momentum at the support
change.

\begin{figure}
        \centering
        \subfigure[]{\includegraphics[width=0.48\textwidth]{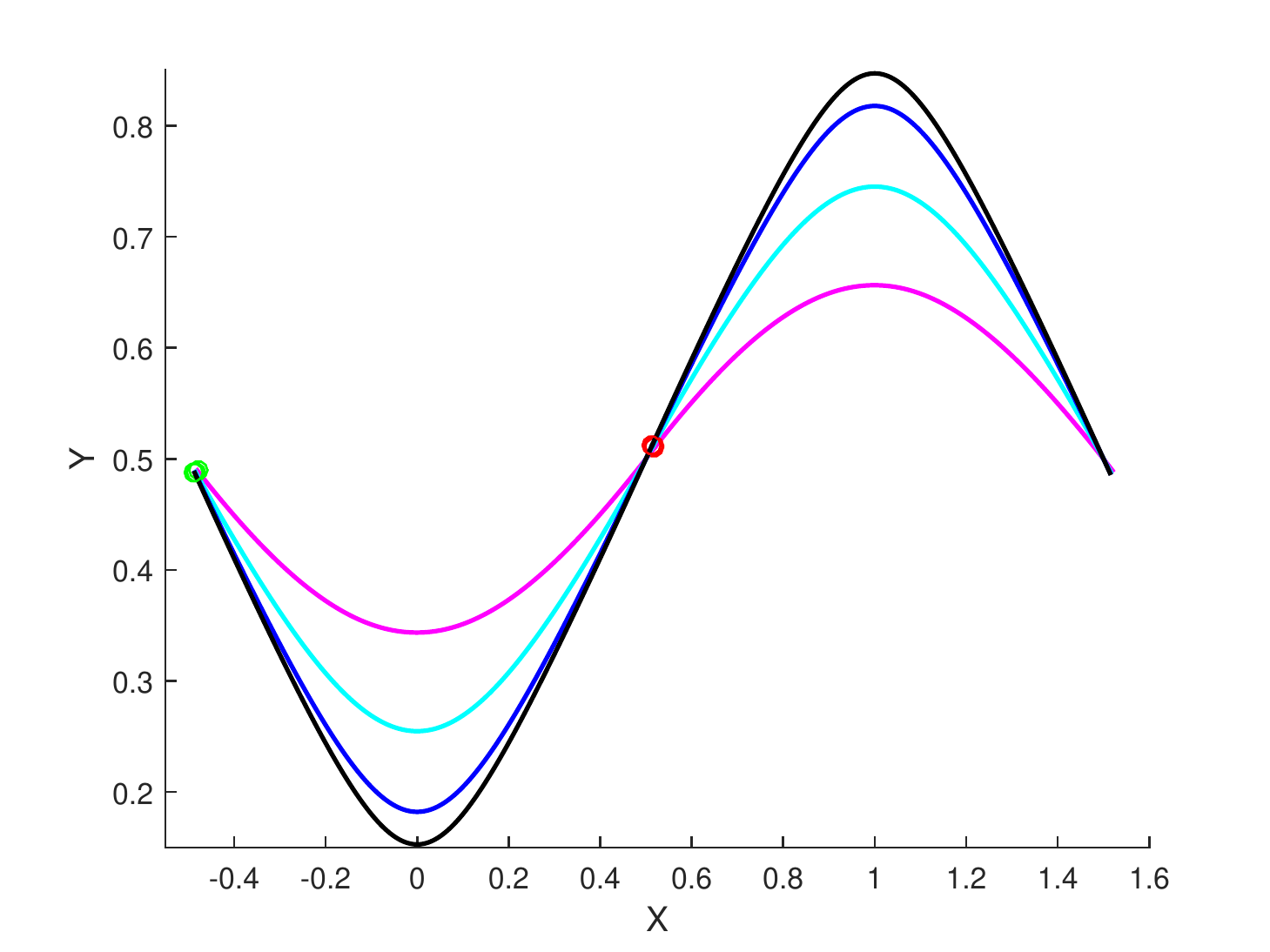}}
        \subfigure[]{\includegraphics[width=0.48\textwidth]{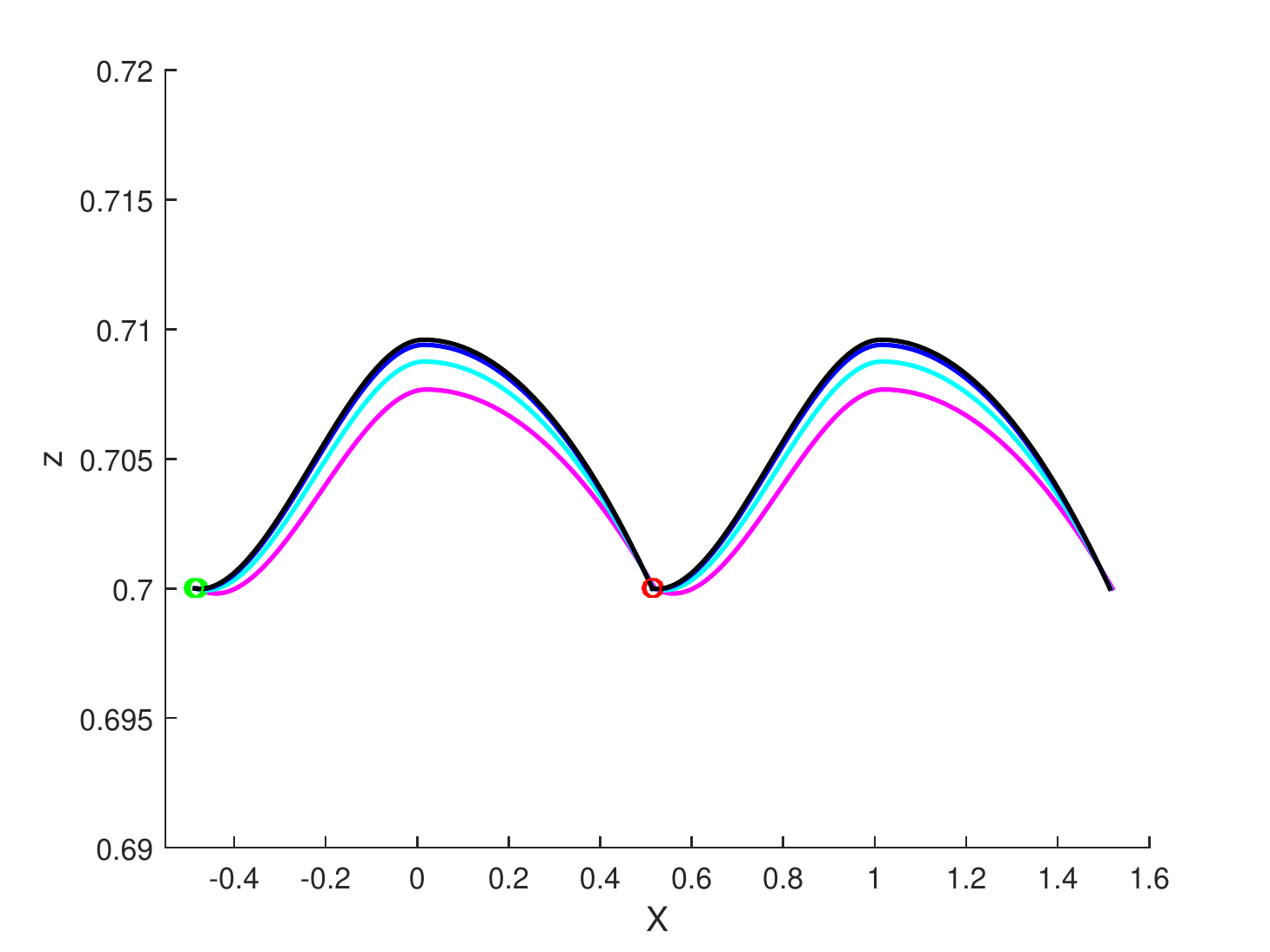}}
        \subfigure[]{\includegraphics[width=0.48\textwidth]{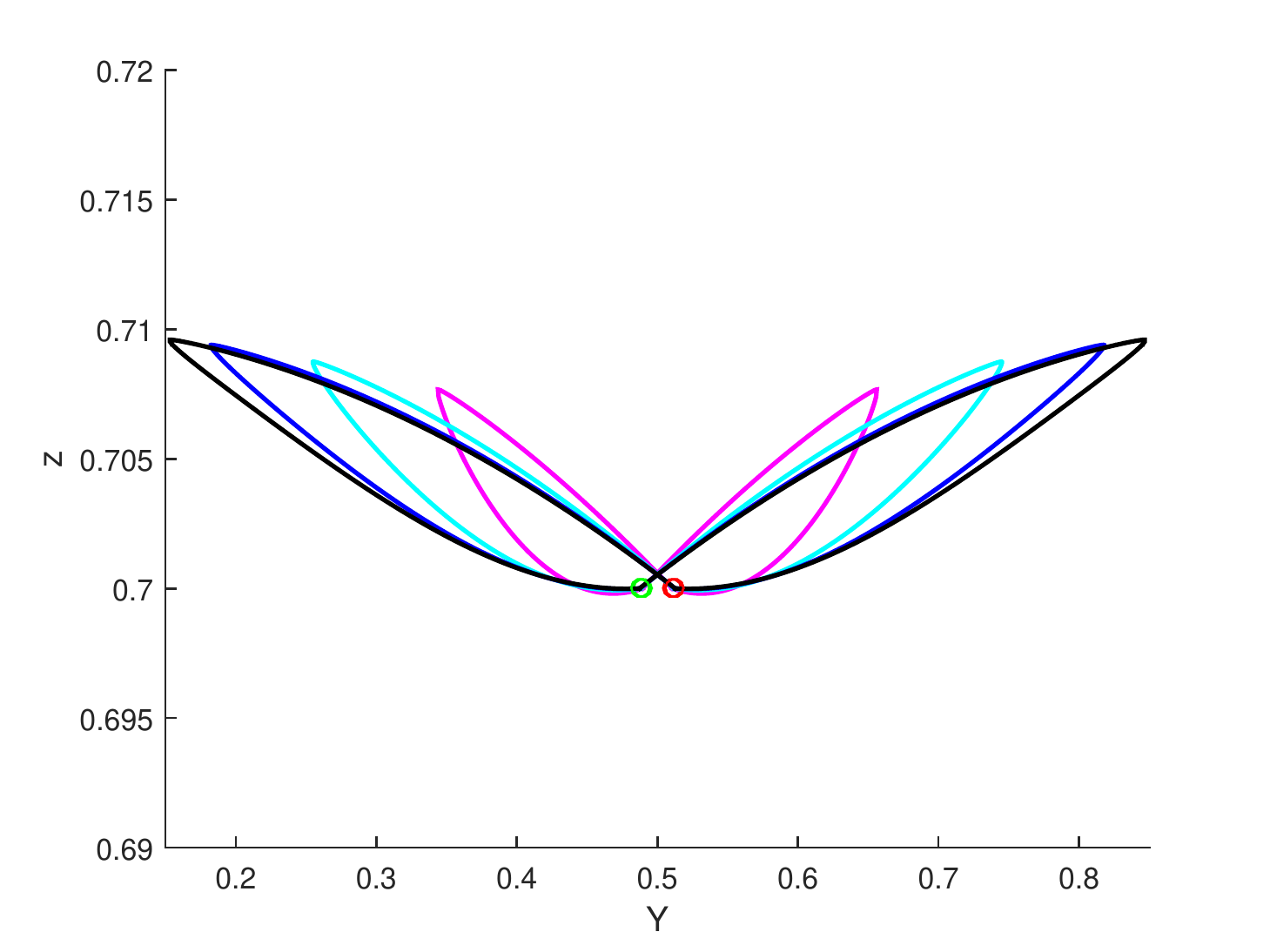}}
        \caption{Motion of the mass in horizontal (a), sagittal (b),
and frontal (c)
planes ($a=0.02$, $C=1.1$, $z_0=0.7$). Magenta, cyan, black, and black
have periods of $T=0.5 \text{ s}, T=0.7 \text{ s}, T=0.9 \text{ s}
$, and  $T=1 \text{ s}$, respectively. Green dots are initial positions and
red dots final positions for one step.}
        \label{fig:motion}
\end{figure}

Figure \ref{fig:motion} illustrates the evolution of the CoM in the
horizontal, sagittal, and frontal planes for several step durations
with $a=0.02 $~m. A figure-eight-shaped
evolution of the CoM can be observed in the frontal
plane, which is qualitatively similar to the motion of the CoM of a human during walking \cite{Saunders53}.

\section{Extension to the control of a humanoid robot model}
\label{sec:humanoid}
We saw in Section \ref{sec:stability_res} that by introducing
oscillations of the CoM and with an appropriate change of support,
it is possible to generate asymptotically stable periodic walking gaits for
the inverted pendulum model. In this section, we apply the same
principles to a more complex model of a bipedal robot. In
particular, we consider a general $n$-DoF humanoid robot
with six joints per leg,  where link mass parameters and
actuator inertias are taken into account.
Numerical simulation will then allow us to discuss the similarities
and differences between this complex model and the simple inverted pendulum with variable leg length which was studied in Section \ref{sec:VLIP-stab}. Indeed, the
simulations will confirm that an asymptotically stable periodic
gait can be achieved for the complex model.

%
%
To illustrate our method of generating stable periodic walking
gaits, we study the robot Romeo \cite{Pateromichelakis14}, \cite[Chapter ~7]
{razavi2016thesis}, which is a 31-DoF humanoid robot constructed by  SoftBank Robotics (https://www.ald.softbankrobotics.com/fr/cool-robots/romeo). Similar to the pendulum model, the
numerical study will focus on the effects of:
\begin{itemize}
        \item the transition condition from step to step as defined by the
 $C$ parameter of the switching manifold defined in
\eqref{eq:switch}; and
        \item vertical oscillations of the CoM as defined by the
parameter $a$.
\end{itemize}
\subsection{A periodic walking gait for the humanoid robot Romeo}


{\color{black} As we would like to reproduce the results obtained for the pendulum, the same virtual constraints used previously and summarized here will be used:
\begin{itemize}
\item The height of the center of mass is given by:
\begin{equation}
\begin{array}{ccc}
  z^d &=& z_0 - a S_a(X,Y) + P(X) ~\mathrm{ for }~ X < DX, \\
  z^d &=& z_0 - a S_a(X,Y) ~\quad\quad\quad\quad\mathrm{ for } ~X \ge DX
\end{array}
\end{equation}
\item The height of the swing foot is given by:
 \begin{equation}
\begin{array}{ccc}
 z_s^d &=& \nu_z S_a(X,Y), + P_z(X) ~\mathrm{ for }~ X < DX, \\
  z_s^d &=& \nu_z S_a(X,Y) ~\quad\quad\quad\quad\mathrm{ for } ~X \ge DX
  \end{array}
\end{equation}
\item The horizontal position of the swing foot is given by:
 \begin{equation}
\begin{array}{lr}
    X_s^d = \left( 1- \nu_X S_a(X,Y)\right) \left(1/2-D_X+X\right) + P_x(X) & \\~\quad\quad\quad\quad\quad\quad\quad\quad\quad\quad\quad\quad\quad\quad\quad\quad\quad\mathrm{ for }~ X < X^l, \\
  X_s^d = \left( 1- \nu_X S_a(X,Y)\right) \left(1/2-D_X+X\right) \mathrm{ for } ~X \ge X^l.\\
     Y_s^d = \left( 1- \nu_Y S_a(X,Y) \right) \left(1/2+D_Y+Y\right) + P_y(X) \\ ~\quad\quad\quad\quad\quad\quad\quad\quad\quad\quad\quad\quad\quad\quad\quad\quad\quad\mathrm{ for }~ X < X^l, \\
  Y_s^d = \left( 1- \nu_Y S_a(X,Y) \right) \left(1/2+D_y+Y\right) ~\mathrm{ for } ~X \ge X^l.
\end{array}
\end{equation}
\end{itemize}
with
 $ S_a(X,Y)=(X-D_x-C D_y)^2+CY^2-((-1/2-C D_y)^2+C(1/2+D_Y^2)$.}

\begin{figure}
\centering
        \includegraphics[width=0.49\textwidth]{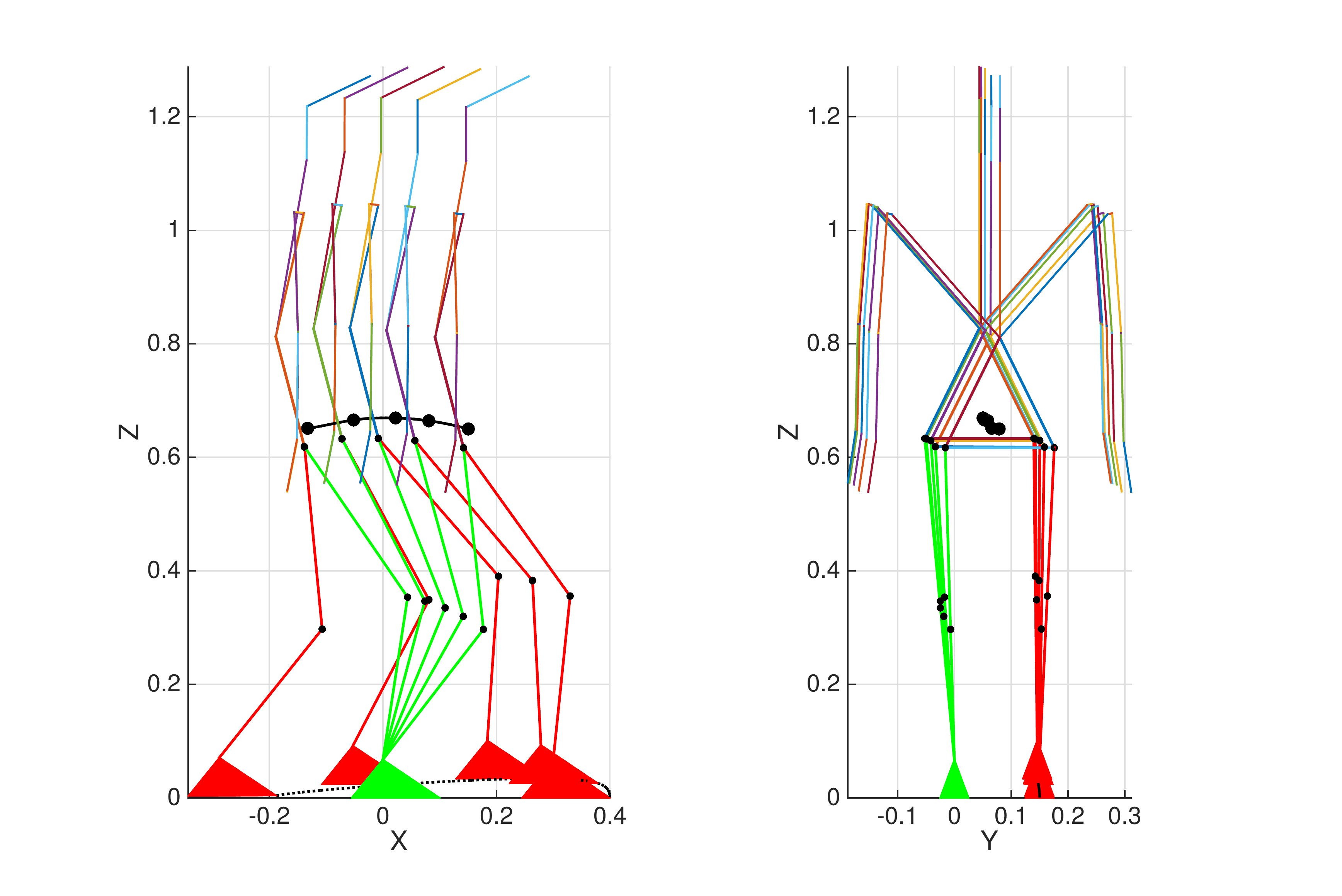}
        \caption{A periodic motion obtained with the proposed
approach for $C=1$, $T=0.55~s$, $a=0.05~m$}
        \label{fig:cyclic_motion}
 \end{figure}

Here, for simplicity,  the remaining DoFs are fixed to realistic constant values\footnote{In practice, we would use optimization to define them.}. The
orientation of the swing foot is controlled to be constant
(the foot remains flat with respect to the ground). The upper-body joints are
constrained to fixed positions and the orientation of the torso is
controlled to be zero (in each of its Euler angles). The fixed upper-body
configuration is visible in Figure \ref{fig:cyclic_motion}.
While the chosen values may have an influence over the gait
stability, examining their potential effects is out of the
scope of this study.

In contrast to the simple pendulum model, it is necessary to take
into account the step length and width in the design of the
constraints due to limits in the robot's workspace and because
the motion of the swing leg affects the dynamics of the robot. Here, we
study a fixed step length $S=0.3~m$ and width $D=0.15~m$.

{\color{black} The periodic motion of the robot is imposed by the parameters defining
the virtual constraints and the HZD model
\eqref{eqn:grizzle:zero_dynamics:reduced_hybrid_model}. Part of the parameters are chosen arbitrarily, and some parameters are defined by optimization to guarantee the existence of a periodic motion compatible with the under-actuation of the robot satisfying some constraints such as duration of the step. Table \ref{table_parameters_chosen} defines the chosen parameters and their effect, while Table \ref{table_parameters_opt} presents the optimized parameters.}

\begin{table}
\begin{tabular}{|c|c|l|}
\hline
  & Value & Effect\\
\hline
a & several & amplitude of oscillations of CoM\\
C & several & shape of the switching manifold \\
T & several & duration of the single support \\
$z_0$ & 0.65~m & minimal high of the CoM \\
$\nu_z$ & 0.09~m (0.045~m) & maximal high of the swing foot \\
$X^l$ & $0.4+D_X$ & limit to adapt the horizontal \\
& & motion of the swing foot \\
\hline
\end{tabular}
\caption{The fixed parameters}
\label{table_parameters_chosen}
\end{table}

\begin{table}
\begin{tabular}{|c|l|}
\hline
  &  Effect\\
\hline
$D_X$ & shift of the CoM position along axis x \\
$D_Y$ & shift of the CoM position along axis y \\
$\nu_x$ & such that the horizontal velocities of \\
$\nu_y$ & the swing leg tip are null at the end of SS\\
\hline
\end{tabular}
\caption{The optimized parameters}
\label{table_parameters_opt}
\end{table}


An example of a walking gait obtained for the robot is illustrated
in Figure \ref{fig:cyclic_motion}. In this example, $C=1$,
$T=0.55~s$, and  $a=0.05~m$. A periodic orbit of the
system is found by calculating a fixed point of the Poincar\'e
return map. Here we have $D_X=0.0016$, $D_Y=0.0259$, $\sigma_X^-
=20.48~kg.m.s^{-1}$, and $\sigma_Y^-=-6.9~kg.m.s^{-1}$.

From the
obtained periodic orbit we observe that the range of motion of the
CoM in the frontal plane is relatively small. Moreover, the obtained
walking gait is highly dynamic as the angular momentum in the
sagittal plane is relatively high, which on the other hand, makes a
small initial velocity insufficient for initiating the walking gait.

We notice that when the walking velocity decreases, the
amplitude of the lateral motion of the CoM increases, and the
workspace limitation for the leg prevents finding a periodic motion for the chosen value of $z_0$.

\label{sec:huma_cyc} 

\subsection{Stability of walking }

\begin{figure}
        \centering
        \psfrag{C}{$C$}
        \psfrag{T}{$T$}
        \includegraphics[width=0.49\textwidth]{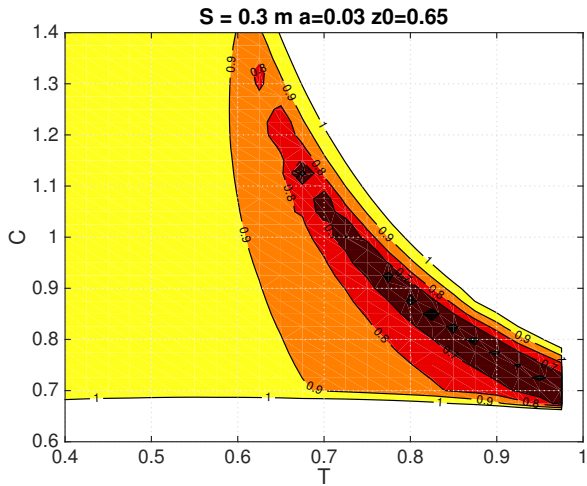}
        \caption{Stability criterion level contours for the
complete model as a function of $C$ and step duration $T$ for
$a=0.03~m$, $S=0.3$~m.}
        \label{fig:complete_A_3}
 \end{figure}

Our main objective is to discover to what extent  the effects of $C$ and $T$ on the
stability of walking for a pendulum model can be extrapolated to the
stability of walking of a realistic humanoid.
Figure \ref{fig:complete_A_3} shows the evolution of the stability
criterion (maximum eigenvalue magnitude of the Jacobian of the
Poincar\'e return map) as a function of $C$ and $T$ for
vertical oscillations with $a=0.03~m$. We observe that the
shape of the stable area as a function of parameters $C$ and $T$ is
quite similar to the one obtained for the pendulum model but with a
shifting of the area of stability toward smaller values of $C$.

\subsection{Comparison between a realistic and a
simplified model}
\begin{figure}
        \centering
                 \psfrag{C}{$C$}
        \psfrag{T}{$T$}
        \includegraphics[width=0.45\textwidth]{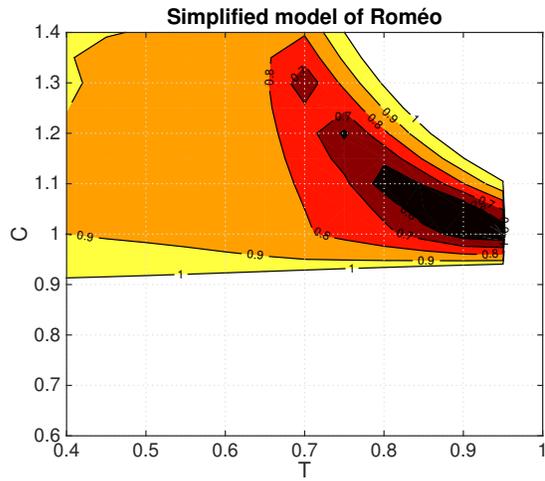}
        \caption{Stability criterion level contours for the
simplified model of Romeo, where a point mass replaces the distributed mass, as
functions of $C$ and step duration $T$ for $a=0.03~m$, $S=0.3~m$. }
        \label{fig:inertia_simp}
 \end{figure}

To properly investigate the effect of having a distributed body for the
trunk and other links versus having a point mass model, as in the
inverted pendulum, we study a simplified model of the
robot by concentrating all the mass of the robot to a point of the trunk.
This point is placed in a position such that with straight legs, the
CoM of the robot and this simplified model coincide. Thus, a model of the robot
with similar kinematics but a different mass distribution is considered. The stable area as a
function of $C$ and $T$  is presented in Figure
\ref{fig:inertia_simp}. We observe that compared to the stability
regions of the realistic model of Romeo, which are shown in  Figure
\ref{fig:complete_A_3}, there is a shifting of the stability areas
along the $C$-axis and a slight modification of the stability margin.
Aside from these differences, the stability regions for the two
models appear quite similar.
As a conclusion, the analysis of the ideal model seems to be an
appropriate tool to understand the key features leading to a stable
gait, but in order to be able to implement the control laws in
practice, studying a more realistic model, like the one presented based on virtual constraints and HZD, is essential to obtain appropriate
numerical values for the key parameters.
\subsection{Effect of the Impact}
\label{sec:impact}
\begin{figure}
        \centering
        \subfigure[]{\includegraphics[width=0.24\textwidth]{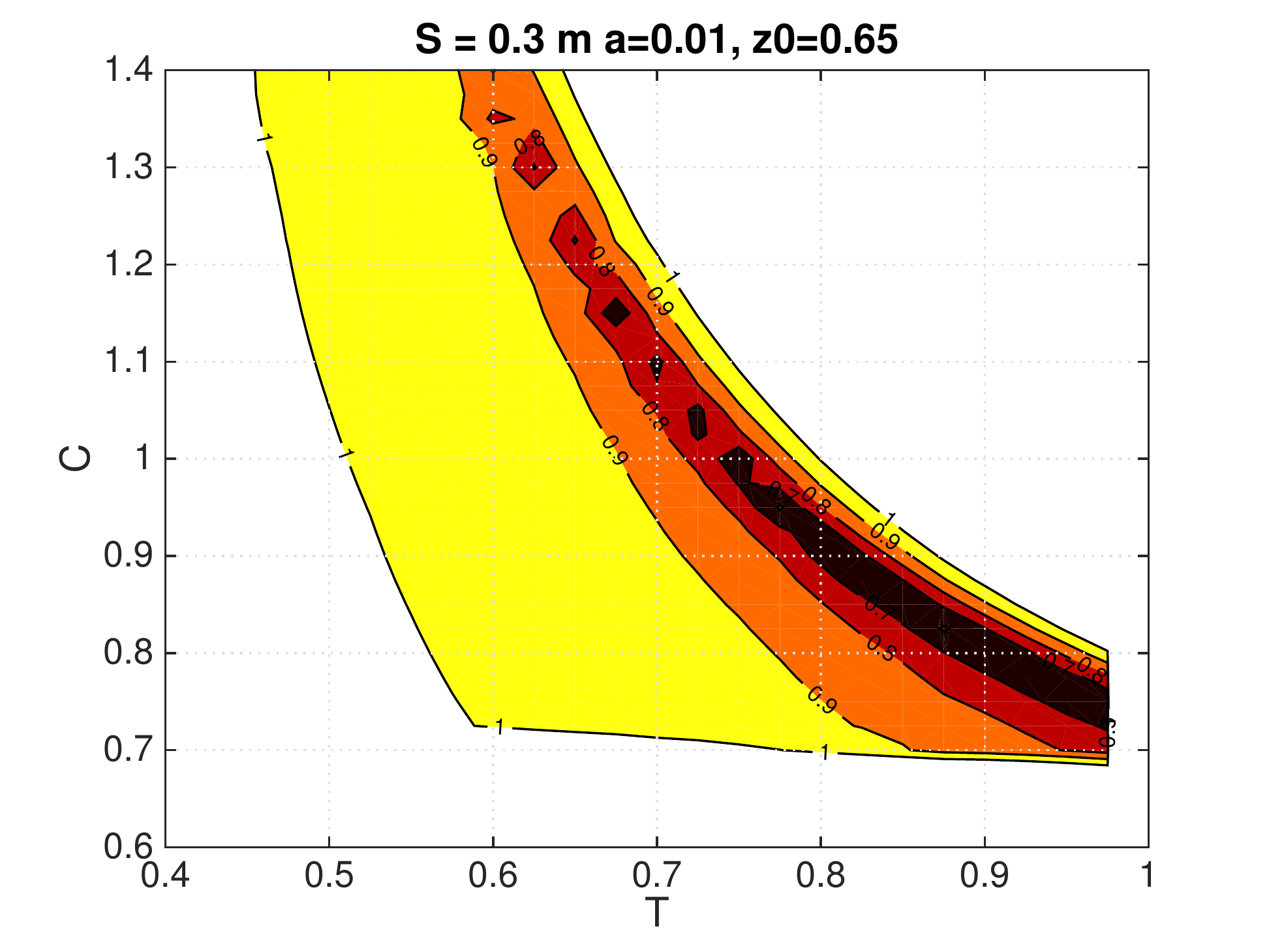}}
        \subfigure[]{\includegraphics[width=0.24\textwidth]{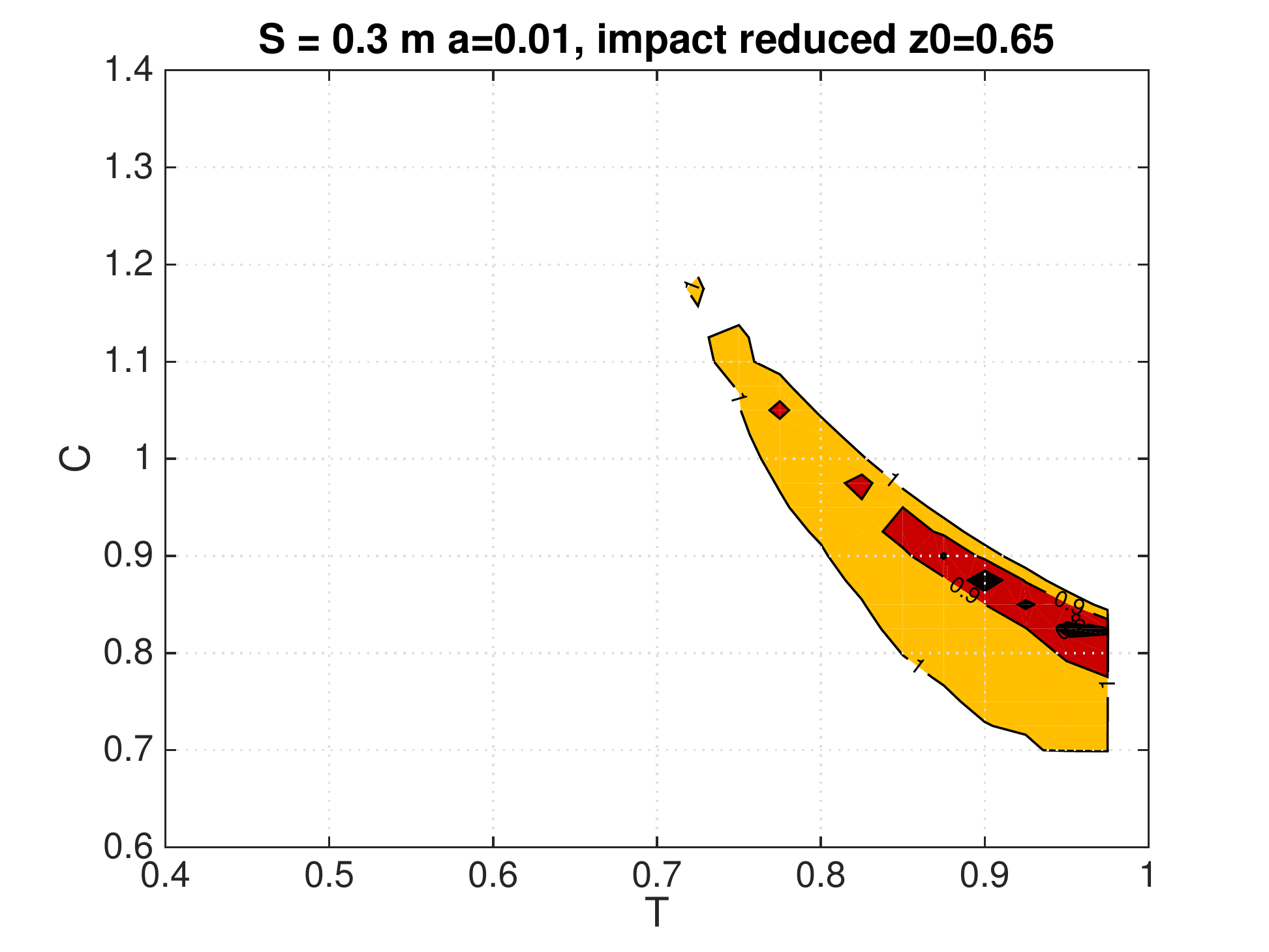}}
\subfigure[]{\includegraphics[width=0.24\textwidth]{\images
Stab_S30_z65_a03_pluspoint.eps}}
\subfigure[]{\includegraphics[width=0.24\textwidth]{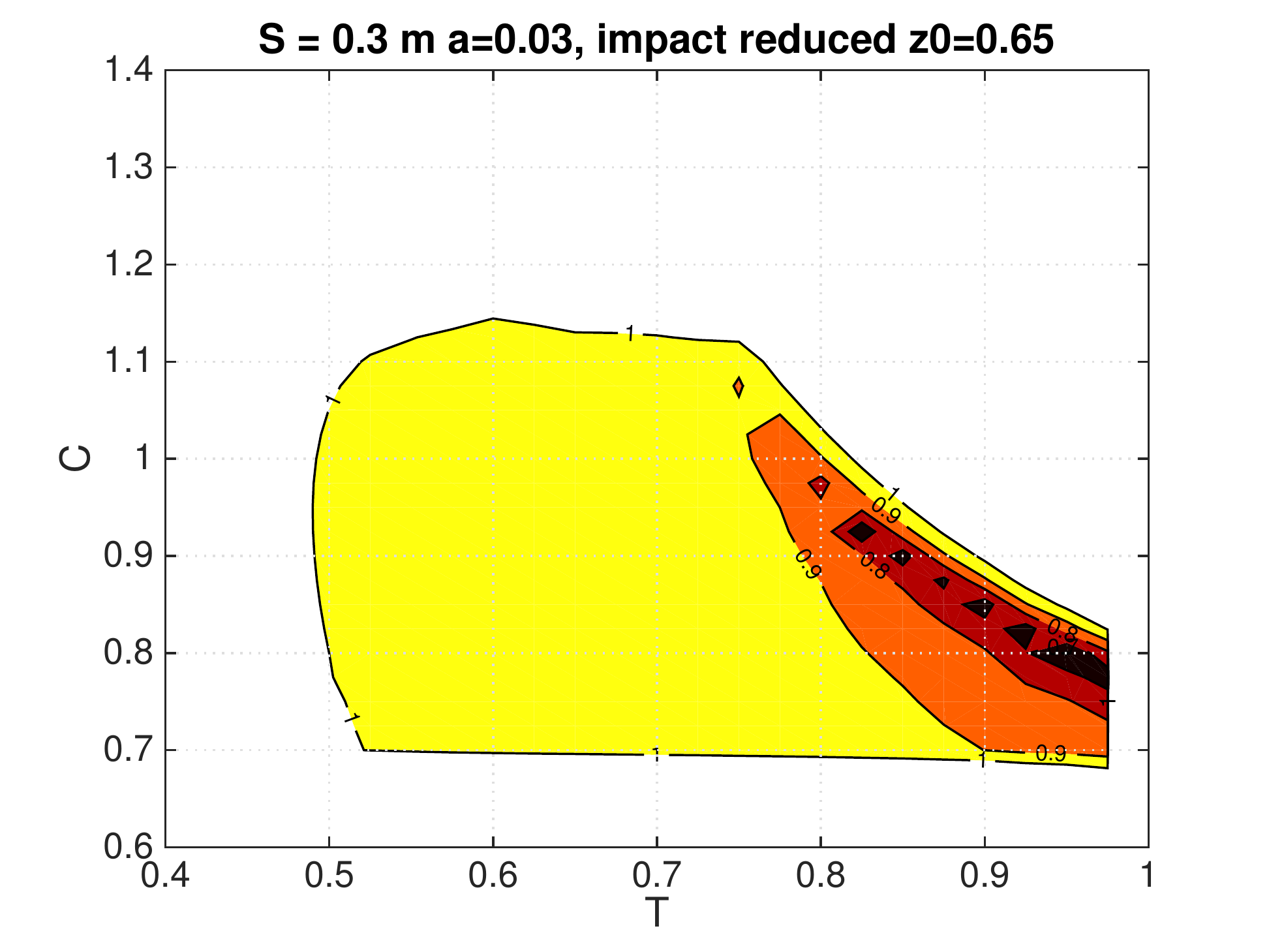}}
\subfigure[]{\includegraphics[width=0.24\textwidth]{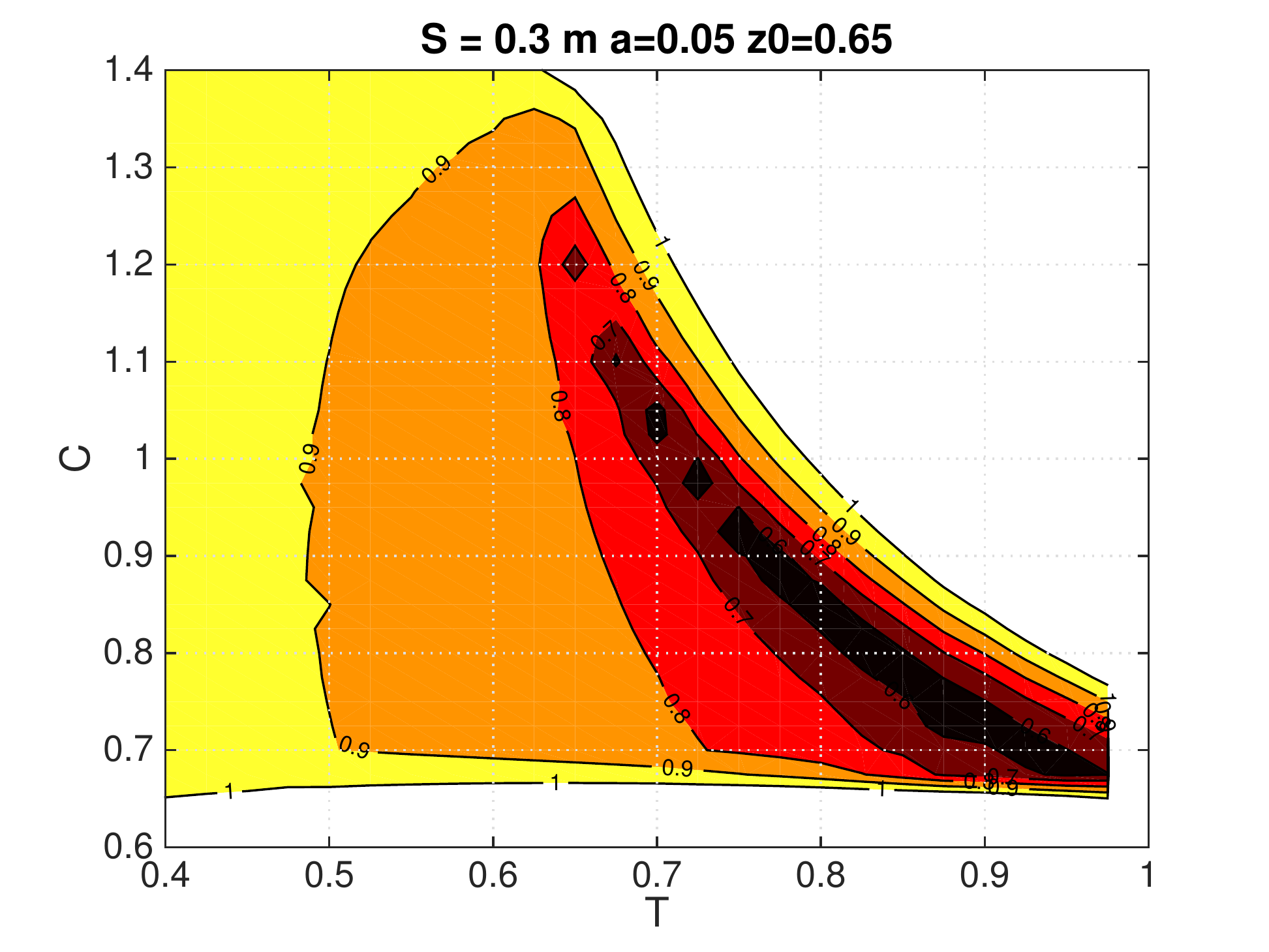}\label{fig:complete_A_5}}
\subfigure[]{\includegraphics[width=0.24\textwidth]{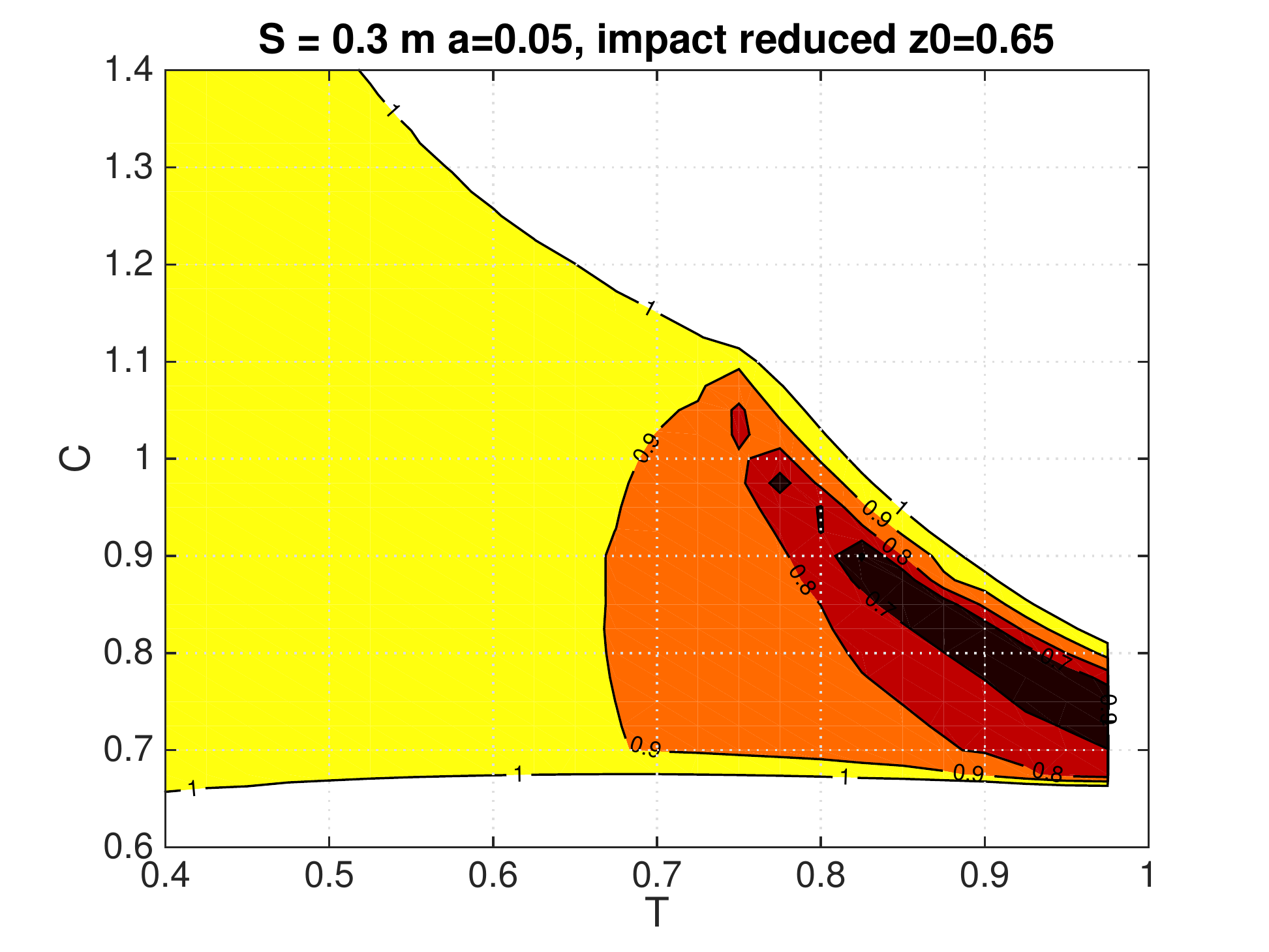}}
        \caption{Stability criterion level contours for the
complete model as function of $C$ and step duration $T$ for  several
amplitudes of oscillations and impacts. For top to down, the
amplitude of oscillations increases $a=0.01~m$; $a=0.03~m$;
$a=0.05~m$. From left to right, the amplitude of impact is divided
by 2.}
        \label{fig:stable_impact_osc}
 \end{figure}

A main difference between the pendulum model and the complete
humanoid model is the impact effect at support change. In the
pendulum model, as the legs are massless, no impact is transmitted
to the CoM when changing the support leg. Thus, the contact of the
swing leg with ground does not affect the pendulum motion. In the
complete robot model, however, legs masses are considered, and a
rigid impact is modeled as the swing leg touches the ground. This
impact has a direct effect on the vertical velocity of the CoM of
the humanoid robot when changing support and thus on the stability
of the gait \cite{Razavi2016}. More specifically, loss of kinetic energy at the
transition between two steps due to the impact or reduction of angular
momentum at the transition \cite{Chevallereau} may result in stable
gaits. To illustrate the contributions of these two mechanisms on
stability, the stable areas as functions of $C$ and $T$ are
demonstrated in Figure \ref{fig:stable_impact_osc}. It can be
clearly seen that an increase of the vertical oscillations via
parameter $a$ leads to an expansion of the stable area and a reduction of the maximal eigenvalue. It can be noted that for a value of
$a=0.05~m$, the real vertical oscillations of the CoM is only around
$0.02~m$ as it can be seen in Figure \ref{fig:robust_COM}. In Figure
\ref{fig:stable_impact_osc}, it is also clear that when the impact
is reduced, via a reduction of the vertical velocity of the swing
leg tip (induced by a reduction of the height of the swing leg
motion right before impact), the area corresponding to the stable
motion shrinks and the maximal eigenvalue increases. In the case
studied, the horizontal velocity of the swing leg tip at impact is
chosen to be null.

\subsection{Simulation of the walking gait}

To illustrate the robustness of the proposed approach with respect
to the configuration and velocity of the robot, this section
presents some different results obtained with the complex model of the humanoid robot.
Starting from rest, the first desired gait corresponds to a walking gait with
$S=0.3~m$, $D=0.15~m$, and $T=0.7~s$ for 30 steps, and then a transition to a  faster
walking gait is imposed with $T=0.55~s$ and the same step sizes for
30 others steps.

\begin{figure}
        \centering
        \includegraphics[width=0.48\textwidth]{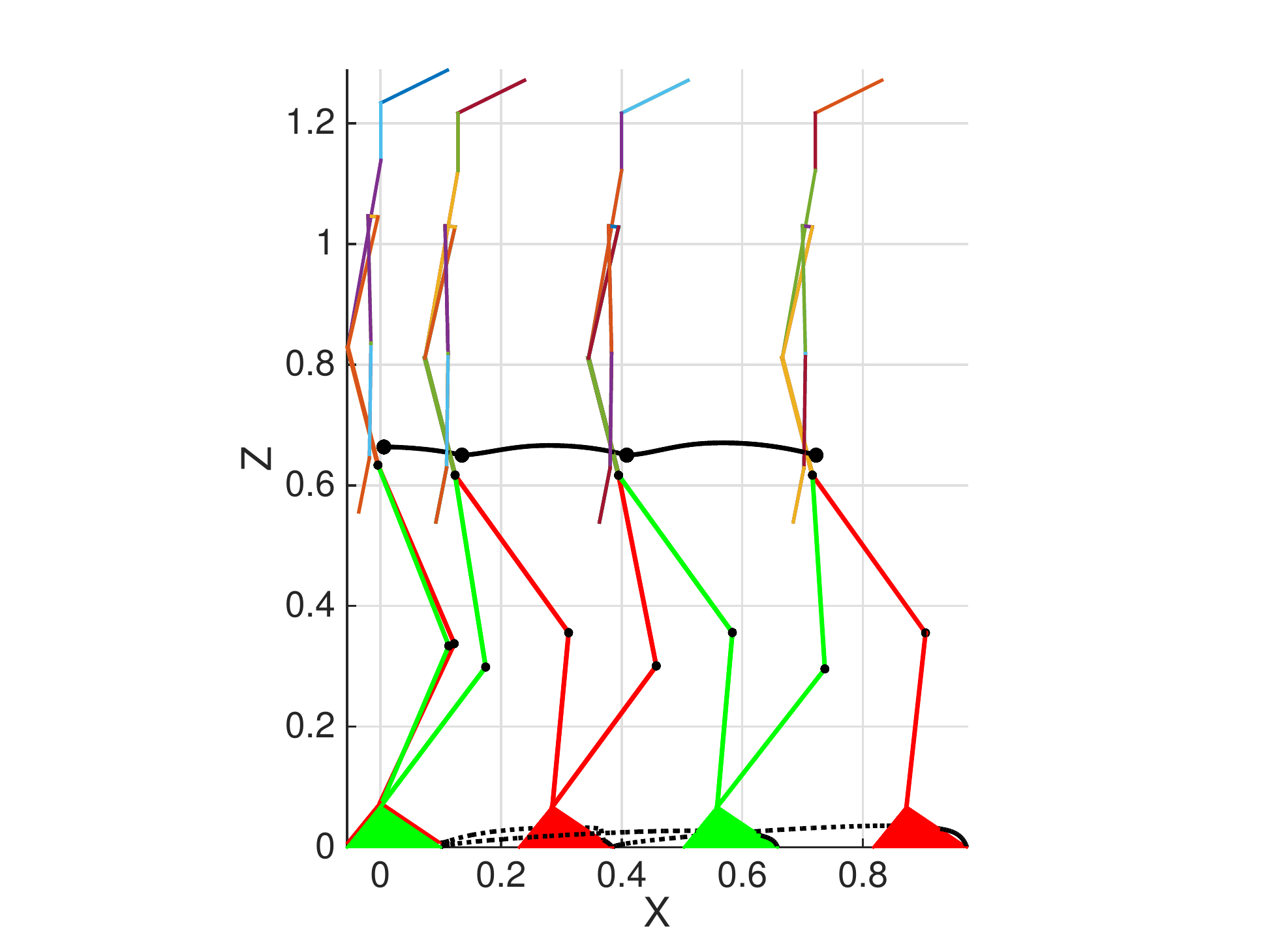}
        \caption{Illustration of the first steps of the ROMEO gait
starting from double support.}
        \label{fig:robust_fig_ini}
\end{figure}

The robot starts its motion in double support with the two legs in
the same frontal plane. In double support, an initial forward motion
of the CoM allows the humanoid to increase its angular momentum around the
axis $\mathbf{n}_0$ such that the barrier of potential energy can be
overcome during the first step. This initial motion places the CoM
in front of the ankle axis and laterally closer to the next stance
foot. This choice of the CoM position at the beginning of the single
support is such that during the first single support phase the
contribution of gravity to the dynamics will be coherent with the
evolution expected during the periodic motion. Based on this initial state,
the virtual constraints are built as presented in Section
\ref{sec:CompVirtCons}. The initial configuration of the robot and a
representation of the first steps in the sagittal plane are shown in
Figure \ref{fig:robust_fig_ini}, where the schematic of the robot is
presented at each transition with the trace of the swing leg tip and
the CoM motions.

\begin{figure}
        \centering
        \includegraphics[width=0.48\textwidth]{\images
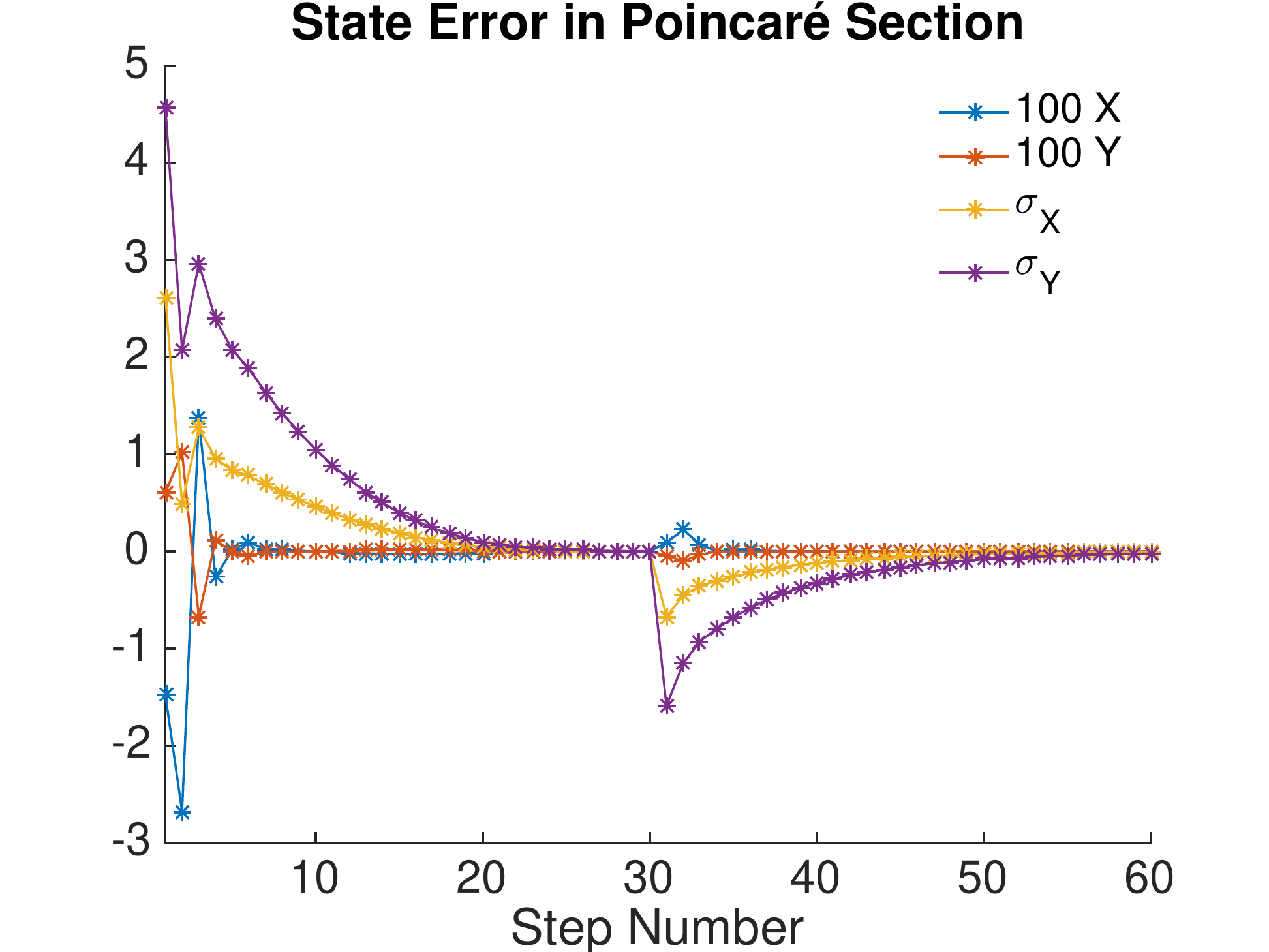}
        \caption{The convergence toward the desired periodic gait is
illustrated via the errors on the unactuated state $X, Y, \sigma_X,
\sigma_Y$ in the Poincar\'e  section as function of the step number.
Due to the scale used, the errors in position are multiplied by
$100$ on the graphic.}
        \label{fig:robust_poincare}
 \end{figure}

 The convergence of the solution to the desired periodic orbit can
be illustrated in various ways. In Figure
\ref{fig:robust_poincare}, the evolution of the state error for the
unactuated variables (defined by the difference between the current
state and the one corresponding to the desired periodic motion) are
shown; the convergence of the error to zero is clear. It can
be observed that the convergence to the periodic motion is faster for
the first motion (with $T=0.7~s$)
compared to the second one ($T=0.55~s$). This result is coherent
with the value of the maximal eigenvalue for the two cases
illustrated in Figure \ref{fig:complete_A_5}.

\begin{figure}
        \centering
        \includegraphics[width=0.48\textwidth]{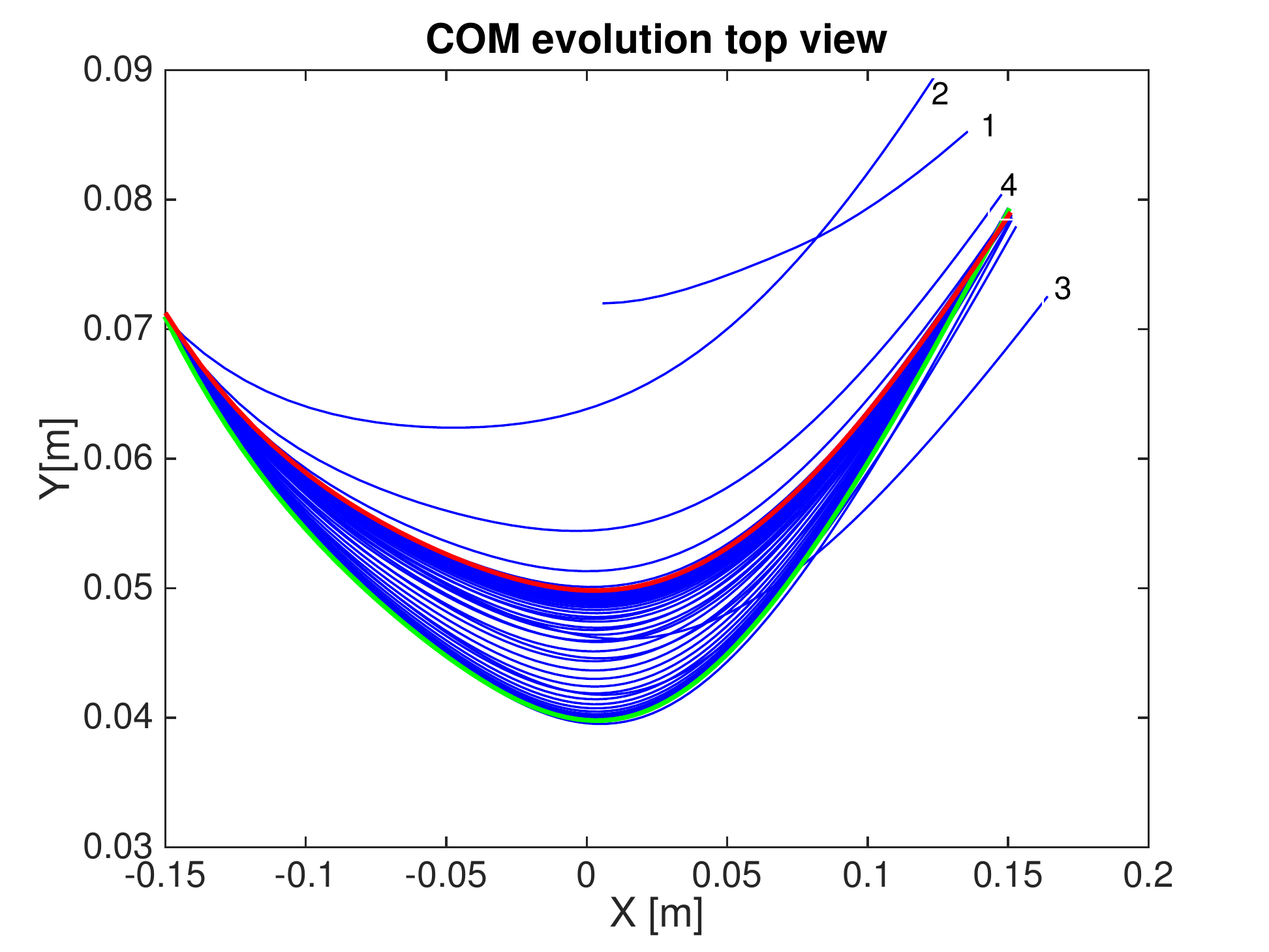}
        \caption{The evolution of the CoM of the robot
in shown in a top view in a frame attached to the stance leg and
lateral axis directed toward the CoM. Each step starts
with the desired initial position of the CoM. The first
steps are numbered, and illustrate a self-synchronization of the
walking gait as observed in \cite{Razavi2016} then the convergence toward
the desired motion is obtained. The green and red curves correspond
respectively to the desired motions with $T=0.7~s$ and $T=0.55~s$.
}
        \label{fig:robust_COM_top}
 \end{figure}

We notice that a faster convergence can be obtained for the error on
position than for the error on angular momentum (or velocity). This
can be interpreted as a fast synchronization between the motion in the
frontal and sagittal planes followed by a stabilization to the
desired motion in terms of velocity. This interpretation is
reinforced by observing the evolution of the CoM in a frame attached
to the stance foot as presented in Figure \ref{fig:robust_COM_top},
where the evolutions corresponding to the first four steps are
numbered. It can be observed that since the swing leg touches the
ground with an imposed distance with respect to the position of the
CoM, the initial position of the CoM at the consecutive steps is
constant (except for the first step). During the first steps a
synchronization is observed due to the choice of the transition
between steps as shown in section \ref{sec:constant}. Then there is a slower
convergence toward the excepted periodic motion while synchronization
is conserved.

We note that even though in the two periodic motions corresponding to
$T=0.7~s$ and $T=0.55~s$ the CoM travels the same distance in the
sagittal plane with $S=0.3~m$, the distances travelled by the CoM in
the frontal plane by the CoM are different.
   \begin{figure}
        \centering
\includegraphics[width=0.48\textwidth]{\images 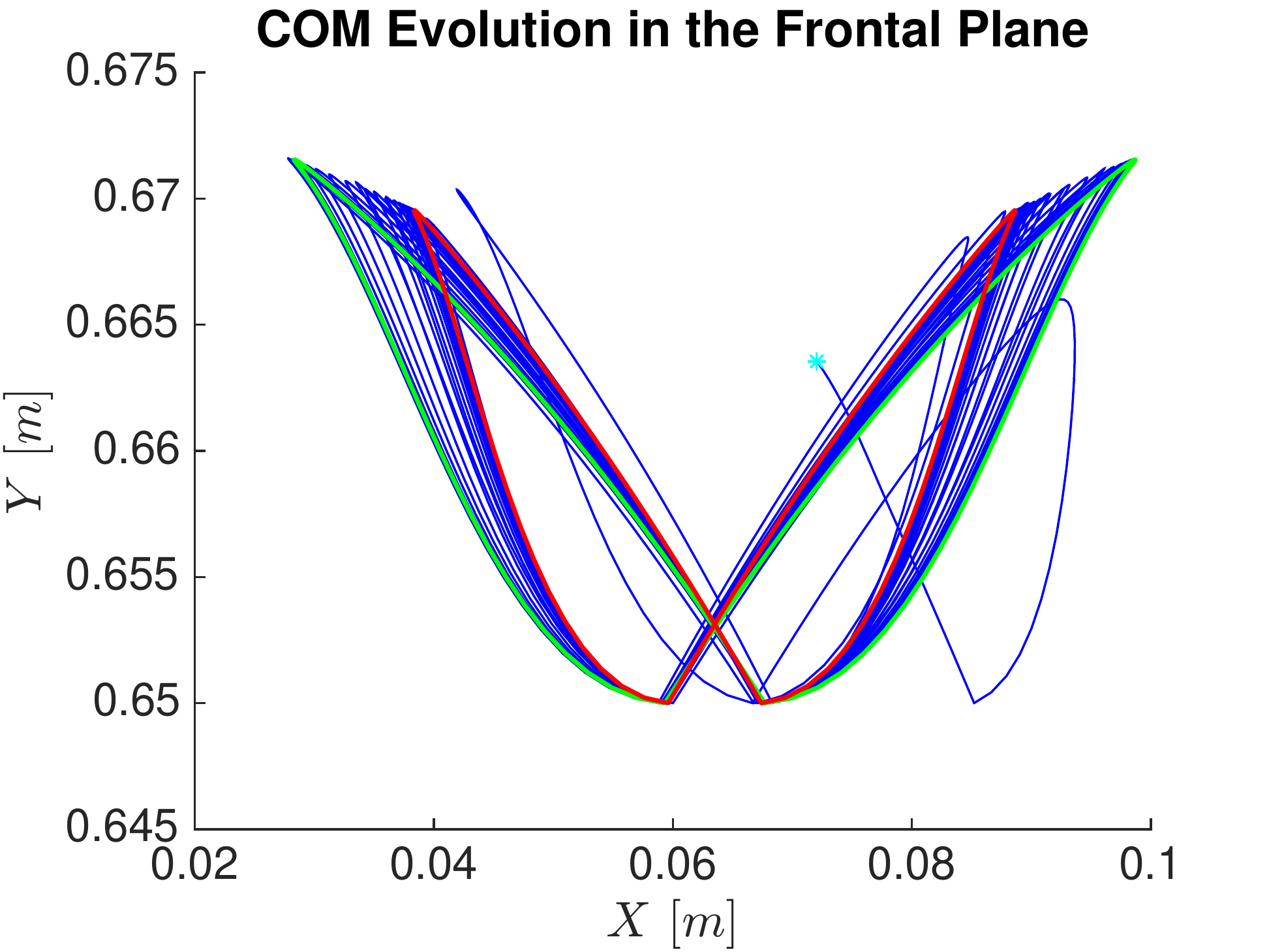}
        \caption{The convergence is shown for the evolution of  CoM in a frontal view in a fixed frame. Starting for the
state denoted by a cyan star, the motion converge toward the green
curve corresponding to the periodic motion with $T=0.7~s$ and then to
the red curve corresponding to periodic motion with $T=0.55~s$.}
        \label{fig:robust_COM}
 \end{figure}

The evolution of the CoM drawn in the frontal plane in a fixed world frame is shown in Figure \ref{fig:robust_COM}. It
can be noticed that the vertical oscillation of the CoM is limited
to approximately $0.02~m$ for $a=0.05~m$; this amplitude
is less than the value observed in human walking.

We can also observe that while the walking gait starts with a foot placed
on the point $(0,0,0)$ in a fixed world frame, and the desired
distance between feet in the frontal plane is $D=0.15~m$, the mean
value of the lateral oscillations of the CoM for the last steps is
different from $D/2$ (see Figure \ref{fig:robust_COM}).
This can be explained by a step width and length that can be
slightly different from $S$ or $D$ during the transient phases. A
correction of the path followed can be easily obtained based on
\cite{ShGrCh2012}.

\begin{figure}
        \centering
        \subfigure[]{\includegraphics[width=0.24\textwidth]
{\images 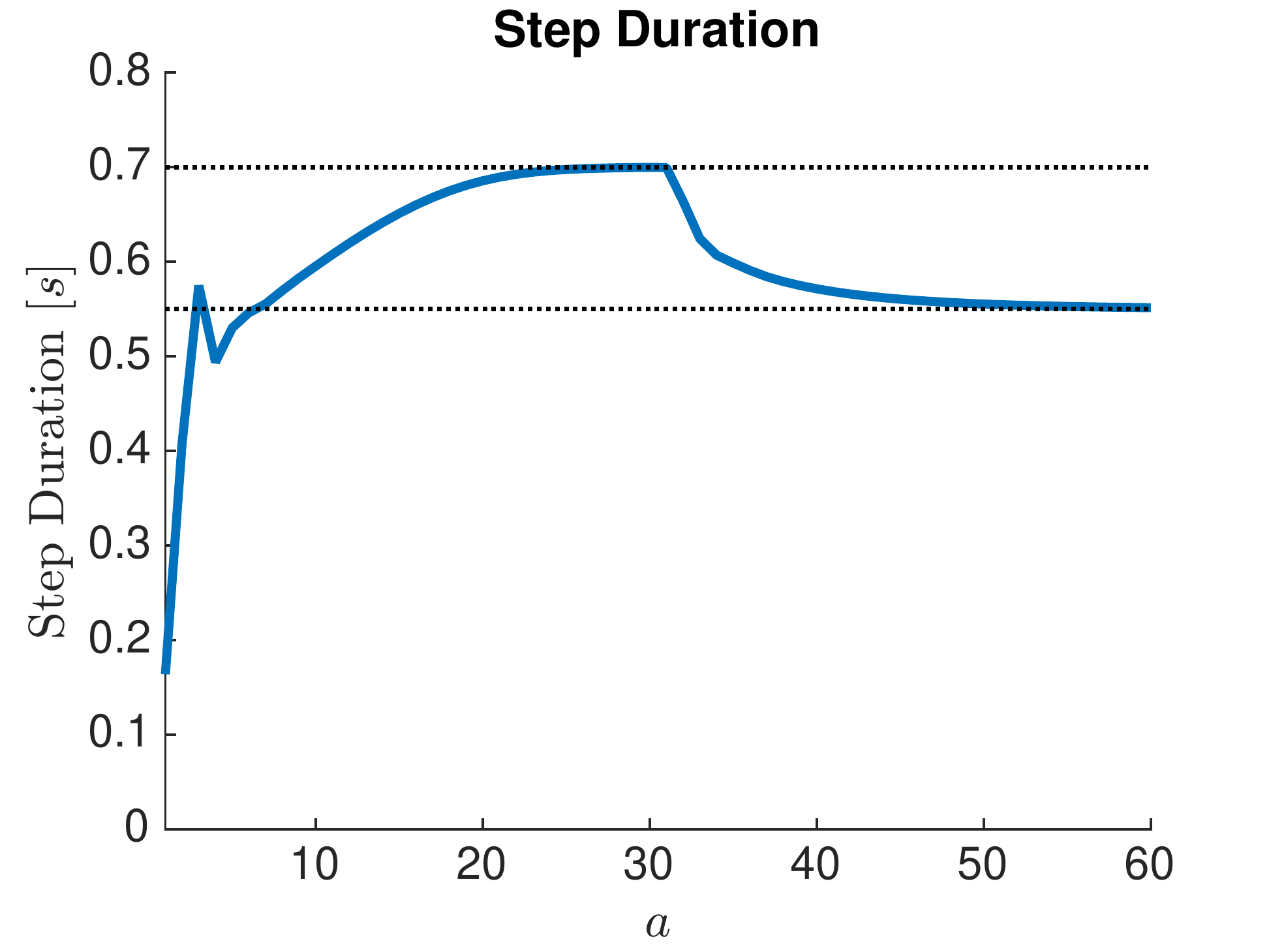}}
 \subfigure[]{\includegraphics[width=0.24\textwidth]{\images
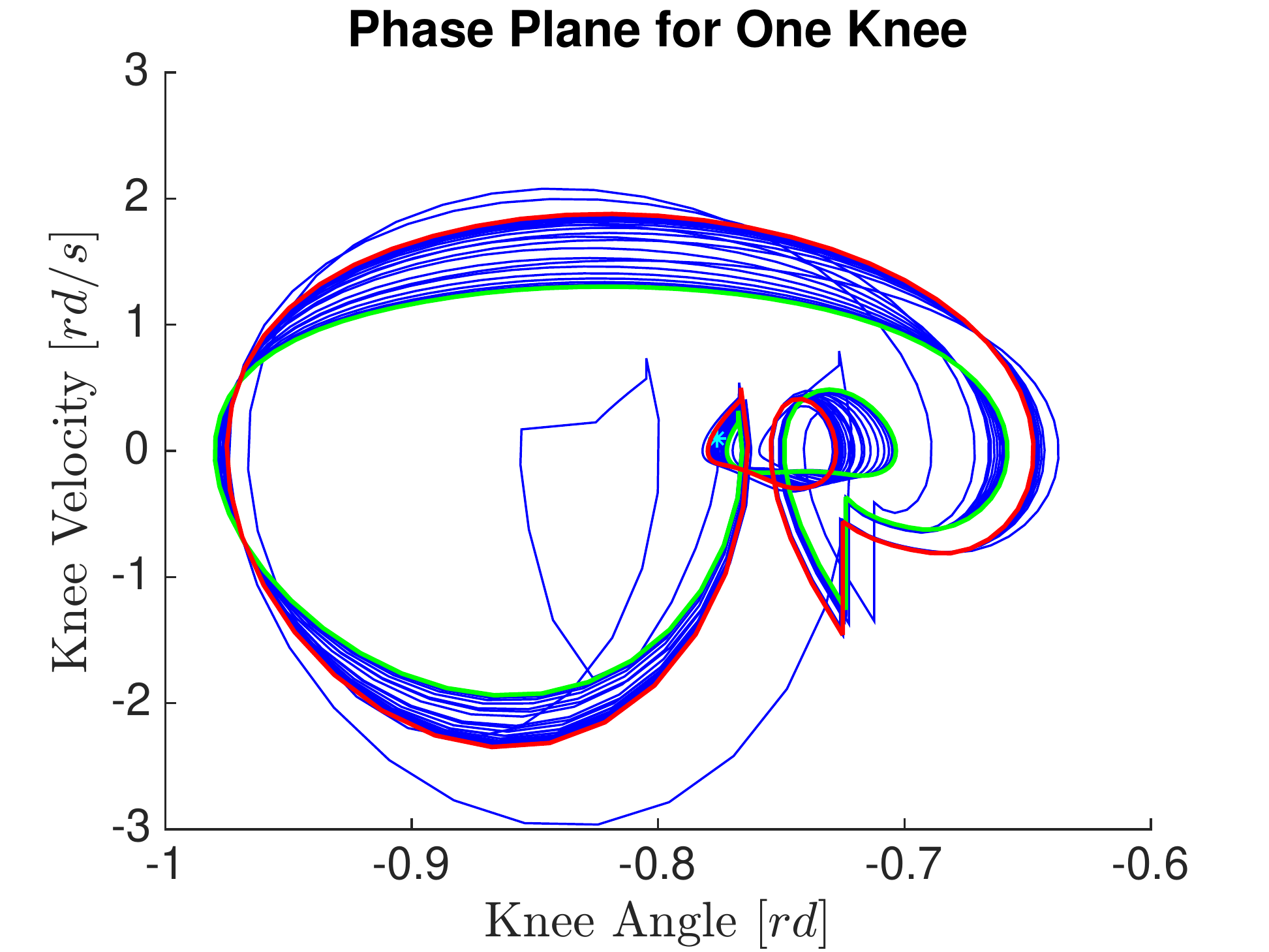}}
        \caption{The convergence is shown for the duration of the
steps as function of the number of steps in (a) and by the evolution
of one knee motion drawn in its phase plane (a). Starting for the
state denoted by a cyan star, the motion converges toward the green
curve corresponding to the periodic motion with $T=0.7~s$ and then to
the red curve corresponding to periodic motion with $T=0.55~s$.}
        \label{fig:robust_knee+T}
 \end{figure}

The convergence toward the desired motion is also illustrated in
Figure \ref{fig:robust_knee+T} for the motion of one joint (the knee)
described in its phase plane and for the evolution of the duration
of the step versus the step number.

\section{Conclusions}

{\color{black} With a control law based on virtual constraints and a planar robot with one degree of underactuation, it is known that the stability of a  walking gait depends only on the periodic orbit itself: if the velocity of the CoM
at the end of the single support phase is directed downward, the gait is stable.
For a 3D humanoid robot with two degrees of underactuation, the condition of stability of a walking gait no longer depends only on the periodic motion but also on the choice of virtual constraints used to create the walking gait. Moreover,
the switching manifold plays a crucial role in the synchronization between motion in the sagittal and frontal planes, and thus on stability.}

This paper studied the influence of the horizontal position of the CoM at the transition from one step to the next and of
vertical oscillations of the CoM within a step on the inherent stability
of a walking gait. Through internal constraints on a simplified
two-legged pendulum model, we were able to obtain asymptotically stable periodic
gaits. The stability properties emerge from the definition of
appropriate virtual constraints. In particular, the walking gait does not
require stabilization through high-level control commands.

We were
able to transfer these properties to a complete robot model and
observe qualitatively similar results.  In doing this, we did encounter new dynamic effects due to energy loss associated with impacts at leg swapping in the complete model which was not considered in the simple pendulum model. It has to be noted that even if the
principles of the approach tested on the simple model can be
transferred to the complex model of the robot, the values of our
key parameters $C$ have to be slightly modified to take into
account the dynamics of the robot. As a consequence, a study of
the Hybrid Zero Dynamics corresponding to the specific robot
considered is useful for obtaining stable walking gaits.

Finally, a complete model of a humanoid has many
degrees of freedom that were not exploited in this study. Here, the upper body and torso
joints have been frozen to arbitrary values. In future work, we
can exploit the upper body joints---and especially the orientation of
the trunk---to increase the energy efficiency of the gaits. Once this is done, our
algorithm will be tested experimentally on the humanoid robot Romeo.

\section*{Appendix}

\subsection*{The control law in single support}

From the definition of the virtual constraints, the output and its
derivatives are
\begin{align}
 y &=  \aq-h_d(\uq),
\label{eqn:grizzle:VC:yformula}\\
  \dot y &= \daq - \frac{\partial h_d
  (\uq)}{\partial \uq} \duq,
\label{eqn:grizzle:VC:dyformula}\\
 \ddot y &= \ddaq - \frac{\partial
h_d(\uq)}{\partial \uq} \dduq -
\frac{\partial}{\partial \uq} \left(\frac{\partial
h_d(\uq)}{\partial \uq} \duq \right)
\duq.
  \label{eqn:grizzle:VC:ddyformula}
\end{align}

The control objective is to ensure that the components of the output vector $y$ converge to zero sufficiently rapidly. Let denote $\nu(y, \dot{y})$  such a control law, yielding
\begin{equation}
\label{eqn:grizzle:VC:eq_closed_loop_a}
   \ddot {y} =  \nu(y, \dot{y}).
\end{equation}
A special case could be $ \nu(y, \dot{y})= -(\frac{K_d}
{\epsilon} \dot{y} +\frac{K_p}{\epsilon^2} y)$ for some $K_p$ and $K_d$ positive definite and $\epsilon >0$ \cite{MoGri2009}.

In the following, we show how one can obtain the control law (i.e., the actuator
torques) ensuring the satisfaction of equation (\ref{eqn:grizzle:VC:eq_closed_loop_a}).

The joint
coordinates, velocities and accelerations of the robot can be
expressed as function of controlled and free variables:
\begin{align}
  q &=  F(\aq,\uq),\label{eqn:grizzle:VC:qformula}\\
  \dot q &=  \frac{\partial F(\aq,\uq)}
{\partial \aq} \daq + \frac{\partial
F(\aq,\uq)}{\partial \uq} \dot q
_f, \label{eqn:grizzle:VC:dqformula} \\
  \ddot q &= \frac{\partial F(\aq,\uq)}
{\partial \aq}  \ddaq + \frac{\partial
F(\aq,\uq)}{\partial \uq}  \dduq
+ \Psi(\aq,\uq, \daq,
\duq),
  \label{eqn:grizzle:VC:ddqformula}
\end{align}
where
\begin{equation}
\Psi(\aq,\uq,\daq,\dot q
_f)=\frac{\partial}{\partial (\aq, \uq) } \left (
\frac{\partial F(\aq,\uq)}{\partial (\aq, \uq) } \left[\begin{array}{l}  \daq \\
\duq \end{array} \right] \right) \left[\begin{array}{l}  \daq \\ \duq
\end{array} \right]. \nonumber
\end{equation}

Using equations \eqref{eqn:grizzle:VC:ddyformula} and
\eqref{eqn:grizzle:VC:eq_closed_loop_a}, it follows that in the
closed-loop system, the accelerations of the controlled variables satisfy
the equation
\begin{equation}
\label{eqn:grizzle:VC:ddqcformula}
\ddaq = \frac{\partial h_d(\uq)}{\partial
\uq} \dduq + \frac{\partial}{\partial
\uq} \left(\frac{\partial h_d(\uq)}{\partial
\uq} \duq \right)\duq+ \nu(y,
\dot{y}),
\end{equation}
where $y$ and $\dot{y}$ are computed in
\eqref{eqn:grizzle:VC:yformula} and
\eqref{eqn:grizzle:VC:dyformula}. The actuator torques able to achieve this desired closed-loop
behavior can be computed based on the dynamic model\eqref{eq:mod_dyn}.
The torque vector $\Gamma$ allowing the $n-2$ controlled
variables to follow the desired closed-loop  behavior
\eqref{eqn:grizzle:VC:ddqcformula}, using \eqref{eqn:grizzle:VC:qformula}, \eqref{eqn:grizzle:VC:dqformula}
\eqref{eqn:grizzle:VC:ddqformula} and \eqref{eq:mod_dyn}, satisfies the
following equation:
\begin{equation}
\label{eqn:grizzle:VC:ModelInNewCoordinates}
\bar{D}(\aq,\uq)
J_r(\aq,\uq)  \dduq +
\Omega(\aq,\uq,\daq,\duq,\nu(y, \dot{y})) = B \Gamma,
\end{equation}
where
\begin{align}\nonumber
\bar{D}(\aq,\uq) &=  D(q) \bigg|_{q=F(\aq,\uq)}, \bigskip \\
J_r(\aq,\uq) &=
  \frac{\partial F(\aq,\uq)}{\partial q
_c} \frac{\partial h_d(\uq)}{\partial \uq}+
\frac{\partial F(\aq,\uq)}{\partial \uq}
\bigskip \nonumber, \\
    \Omega &= \bar{D}(\aq,\uq
\bigg[  \frac{\partial F(\aq,\uq)}{\partial
\aq}\\   & \bigg( \frac{\partial}{\partial \uq}
  \left(\frac{\partial h_d(\uq)}{\partial \uq}
\duq \right) \duq+ \bigskip \nonumber
\nu(y, \dot{y})
\bigg) \\ & + \Psi(\aq,\uq,\daq,
\duq)
\bigg] + \bar{H}(\aq,\uq,\dot q
_c,\duq), \bigskip \nonumber \\
\bar{H}(\aq,\uq,\daq ,
\duq)&=  H(q,\dot{q}) \bigg|_{\begin{array}{l}
q=F(\aq,\uq)\\ \dot q
=\frac{\partial F(\aq,\uq)}{\partial \aq}
\daq + \frac{\partial F(\aq,\uq)}
{\partial \uq} \duq \end{array} } \nonumber
\end{align}
Multiplying \eqref{eqn:grizzle:VC:ModelInNewCoordinates} on the left
by the full rank matrix
$$\left[
      \begin{array}{c}
        B^\perp \\
        B^+
      \end{array}\right],
      $$
where $B^+$ is the pseudo inverse of $B$ and $
B^\perp$ is a $2 \times n$ matrix of rank $2$ such that $
B^\perp B=0$, results in
\begin{align}
\label{eq:HZD_Borth}
        B^\perp \bar{D}(\aq,\uq)
 J_r(\aq,\duq) \dduq +
B^\perp \Omega(\aq,\uq,\daq,
\duq,y,\dot{y})
&= 0 \\
        B^+ \bar{D}(\aq,\uq)
J_r(\aq,\uq)
\dduq +B^+ \Omega(\aq,\uq,
\daq,\duq,y,\dot{y}) &=
\Gamma.
\end{align}
It follows that a feedback control law ensuring
\eqref{eqn:grizzle:VC:eq_closed_loop_a} can be obtained from the
following equations
\begin{align}
\label {eqn:grizzle:VC:FeedbackEfficient}
\Gamma&= B^+ \bar{D}(\aq,
\uq)   J_r(\aq,\uq) v +B^+
\Omega(\aq,\uq,\daq,\duq,y,\dot{y}) \\
 v &= -\left(B^\perp \bar{D}(\aq,
\uq
 J_r(\aq,\uq)\right)^{-1} B
^\perp \Omega(\aq,\uq,\daq,
\duq,y,\dot{y}) \label{eq:appendix}.
\end{align}
{\color{black}where $v$ denoted the acceleration $\dduq$ satisfying the equation \eqref{eq:HZD_Borth}.}
This form of the feedback law only requires the inversion of a
matrix whose size corresponds to the number of unactuated
coordinates, here a 2$\times$2 matrix.

\subsection*{Condition of synchronization for the LIP model}
{\color{black} This section details of the condition of synchronization for the LIP model and the proposed control strategy, with the virtual constraints proposed in section \ref{sec:CompVirtCons}.}

{\color{black} The initial state of the robot after step {i} is written as:
\begin{eqnarray}
X^+_i &=& -1/2, \label{eq:state1} \\
Y^+_i &=&  1/2 \\
\dot X^+_i &=& \dot X_0 + \delta \dot{X}^+_i \\
\dot Y^+_i &=& \dot Y_0 + \delta \dot{Y}^+_i
\end{eqnarray}
Due to the discrete invariance of the control law, there is no error on position.
At the end of the step the state of the robot will be denoted
\begin{eqnarray}
X^-_i &=&  1/2 + \delta {X}^-_i \\
Y^-_i &=&  1/2 + \delta {Y}^-_i \\
\dot X^-_i &=& \dot X_0 + \delta \dot{X}^-_i \\
\dot Y^-_i &=& -\dot Y_0 + \delta \dot{Y}^-_i
\label{eq:state6}
\end{eqnarray}}

{\color{black} Using the fact that $(\dot X_0$ and $\dot Y_0)$ and $L_i=0$ define a synchronized motion and neglecting the second order terms, we obtain:
\begin{eqnarray}
L_i\thickapprox \dot X_0 \delta \dot{Y}^+_i + \dot Y_0  \delta \dot{X}^+_i \label{eq:cons_sync1}
\end{eqnarray}
Since the velocity of the CoM is conserved through change of support for the LIP model:
\begin{eqnarray*}
\delta \dot X^+_{i+1} &=&  \delta \dot{X}^-_i \\
\delta \dot Y^+_{i+1} &=& - \delta \dot{Y}^-_i
\end{eqnarray*}
and $L_{i+1}$ can be expressed as
\begin{equation}
L_{i+1}\thickapprox \dot X_0 \delta \dot{Y}^+_{i+1} + \dot Y_0  \delta \dot{X}^+_{i+1}
\end{equation}
or
\begin{equation}
L_{i+1}\thickapprox- \dot X_0 \delta \dot{Y}^-_{i} + \dot Y_0  \delta \dot{X}^-_{i}
\end{equation}
\begin{eqnarray}
\frac{L_{i+1}}{L_i}\thickapprox \frac{- \dot X_0 \delta \dot{Y}^-_{i} + \dot Y_0  \delta \dot{X}^-_{i} }{\dot X_0 \delta \dot{Y}^+_{i} + \dot Y_0  \delta \dot{X}^+_{i} }
\label{eq:ratio_L}
\end{eqnarray}
We will now express the final error in velocity as function of the initial error for the step $i$.
The orbital energies, $\mathcal{E}_x$ and $\mathcal{E}_y$ and synchronization measure $L$ are conserved quantities. Therefore, we have,
\begin{eqnarray}
\left(\dot{X}_i^- \right)^2-\omega^2 \left(X^-_i\right)^2 &=& \left(\dot{X}_i^+\right)^2-\omega^2 \left( X^+_i \right) ^2\\
\left(\dot{Y}_i^- \right)^2-\omega^2 \left(Y^-_i\right)^2 &=& \left(\dot{Y}_i^+\right)^2-\omega^2 \left( Y^+_i \right) ^2\\
\dot{X}_i^- \dot{Y}_i^- - \omega^2 X^-_i Y^-_i&=&L_i
\end{eqnarray}
or
\begin{eqnarray*}
\left(\dot X_0 + \delta \dot{X}^-_i \right)^2-\omega^2 \left(1/2 + \delta {X}^-_i \right)^2 \\ = \left(\dot X_0 + \delta \dot{X}^+_i \right)^2-\omega^2/4 \\
\left(-\dot Y_0 + \delta \dot{Y}^-_i \right)^2-\omega^2 \left(1/2 + \delta {Y}^-_i \right)^2 \\ = \left(\dot Y_0 + \delta \dot{Y}^+_i \right)^2-\omega^2/4
\\
\left( \dot X_0 + \delta \dot{X}^-_i \right) \left( -\dot Y_0 + \delta \dot{Y}^-_i \right)\\ - \omega^2 \left( 1/2 + \delta {X}^-_i \right) \left( 1/2 + \delta {Y}^-_i \right) = L_i.
\end{eqnarray*}
Using these equations and neglecting the second order terms we obtain:
\begin{eqnarray}
 \delta \dot{X}^-_i  &\thickapprox& \frac{ \omega^2}{2 \dot X_0}  \delta {X}^-_i + \delta \dot{X}^+_i \label{eq:cons_ener_x} \\
 \delta \dot{Y}^-_i  &\thickapprox& -\frac{\omega^2}{2 \dot Y_0}  \delta {Y}^-_i - \delta \dot{Y}^+_i \label{eq:cons_ener_y}
 \\
\frac{\omega^2}{2} \left( \delta {X}^-_i  + \delta {Y}^-_i \right)
&\thickapprox& \dot X_0 \delta \dot{Y}^-_i - \dot Y_0 \delta \dot{X}^-_i - L_i.\nonumber \\
\label{eq:cons_L_annexe}
\end{eqnarray}
The last equation \eqref{eq:cons_L_annexe} combines with the previous ones  \eqref{eq:cons_ener_x} \eqref{eq:cons_ener_y} can give us a relation between $\delta {X}^-_i$, $\delta {Y}^-_i$ and $L_i$.
\begin{eqnarray}
\frac{\omega^2}{2} \left( \delta {X}^-_i  + \delta {Y}^-_i \right)
\thickapprox  -\frac{\omega^2 \dot X_0}{2 \dot Y_0}  \delta {Y}^-_i  -  \frac{ \omega^2 \dot Y_0}{2 \dot X_0}  \delta {X}^-_i - 2 L_i
\end{eqnarray}
\begin{equation}
\left( 1 + \frac{  \dot Y_0}{ \dot X_0}\right) \delta {X}^-_i  + \left( 1 + \frac{ \dot X_0}{ \dot Y_0} \right) \delta {Y}^-_i
\thickapprox  - 4 \frac{L_i}{\omega^2}.
\label{eq_err_pos L}
\end{equation}
Equation \eqref{eq_err_pos L} confirms that synchronized motion occurs with $\delta {X}^-_i=\delta {Y}^-_i=0$ and in this case via \eqref{eq:cons_ener_x} and \eqref{eq:cons_ener_y}, we see that the error on velocity is conserved.
Now we will write the ratio $\dfrac{L_{i+1}}{L_{i}}$ as function of $C$.
The change of support occurs on the switching manifold \eqref{eq:switch2}. For small variation around the periodic motion, the final position error for each step satisfies:
\begin{equation}
\delta X^-_i = - C \delta Y^-_i.
\end{equation}
Using \eqref{eq_err_pos L}, the error in position at the end of the step, can deduced as function of $L_i$. We have:
\begin{equation}
\delta {Y}^-_i \thickapprox  - 4 \frac{L_i \dot X_0 \dot Y_0}{\omega^2 (-\dot Y_0 C + \dot X_0)(\dot X_0+\dot Y_0)}.
\end{equation}
The others errors can also be deduced:
\begin{equation}
\delta {X}^-_i\thickapprox   4 C \frac{L_i \dot X_0 \dot Y_0}{\omega^2 (-\dot Y_0 C + \dot X_0)(\dot X_0+\dot Y_0)}
\end{equation}
\begin{equation}
\delta \dot{X}^-_i  \thickapprox 2 C \frac{L_i \dot X_0 \dot Y_0}{\dot X_0 (-\dot Y_0 C + \dot X_0)(\dot X_0+\dot Y_0)} + \delta \dot{X}^+_i
\end{equation}
\begin{equation}
 \delta \dot{Y}^-_i  \thickapprox  2 \frac{L_i \dot X_0 \dot Y_0}{\dot Y_0 (-\dot Y_0 C + \dot X_0)(\dot X_0+\dot Y_0)} - \delta \dot{Y}^+_i
\end{equation}
We can then calculate the change of synchronization value step by step, we obtain:
\begin{equation}
\frac{L_{i+1}}{L_i}\thickapprox \frac{- \dot X_0 \delta \dot{Y}^-_{i} + \dot Y_0  \delta \dot{X}^-_{i} }{L_i}\label{eq:LplusisurLi}
\end{equation}
Using \eqref{eq:cons_ener_x} and \eqref{eq:cons_ener_y} equation \eqref{eq:LplusisurLi} becomes:
\begin{eqnarray}
\frac{L_{i+1}}{L_i}\thickapprox \frac{ \dot X_0 \delta \dot{Y}^+_{i} + \dot Y_0  \delta \dot{X}^+_{i} }{L_i} +   \nonumber\\ 2 (C \frac{\dot Y_0}{\dot X_0}-\frac{\dot X_0}{\dot Y_0} ) \frac{\dot X_0 \dot Y_0}{(-\dot Y_0 C + \dot X_0)(\dot X_0+\dot Y_0)}\label{eq:LplusisurLi1}
\end{eqnarray}
With \eqref{eq:cons_sync1} equation can be rewritten \eqref{eq:LplusisurLi1} as follows:
\begin{equation}
\frac{L_{i+1}}{L_i}\thickapprox  \frac{(\dot Y_0-\dot X_0)( C \dot Y_0 +\dot X_0)}{(\dot X_0+\dot Y_0)( -C \dot Y_0 + \dot X_0)}\label{eq:Liplus1overLi}
\end{equation}}

\section{Acknowledgments}
We gratefully acknowledge partial support by ANR Equipex Robotex. The work of J. Grizzle was supported by NSF Awards NRI-1525006 and Inspire-1343720.

\bibliographystyle{plain}
\bibliography{Articlemaster}
\end{document}

%% file: Romeo_VC_headerFile.tex
\newcommand{\Q}{{\cal Q}}

\newcommand{\aq}{q_{\rm c}}
\newcommand{\daq}{\dot{q}_{\rm c}}
\newcommand{\ddaq}{\ddot{q}_{\rm c}}

\newcommand{\ulbl}{\rm f}
\newcommand{\uq}{q_{\ulbl}}
\newcommand{\duq}{\dot{q}_{\ulbl}}
\newcommand{\dduq}{\ddot{q}_{\ulbl}} 

\newcommand{\zerolbl}{\rm zero}
\newcommand{\zdmanifold}{\mathcal{Z}}
\newcommand{\zero}{{\rm zero}}

\newcommand{\qfree}{q_{\rm f}}

\newcommand{\usigma}{\sigma_{\ulbl}}
\newcommand{\dusigma}{\dot{\sigma}_{\ulbl}}

\newcommand{\impactmap}{\varDelta}
\newcommand{\impactmappos}{\varDelta_{q}}
\newcommand{\impactmapvel}{\varDelta_{\dot q}}

\newcommand{\fzero}{f_{\zerolbl}}

\newcommand{\impactmapzero} [1][] {\impactmap_{{\zerolbl}#1}}